\documentclass[letterpaper]{article} %
\usepackage{aaai23}  %
\usepackage{times}  %
\usepackage{helvet}  %
\usepackage{courier}  %
\usepackage[hyphens]{url}  %
\usepackage{graphicx} %
\usepackage{natbib}  %
\usepackage{caption} %
\usepackage{algorithm}
\usepackage{newfloat}
\usepackage{listings}
\DeclareCaptionStyle{ruled}{labelfont=normalfont,labelsep=colon,strut=off} %
\floatstyle{ruled}
\newfloat{listing}{tb}{lst}{}
\floatname{listing}{Listing}
\usepackage{amsmath,amsfonts,bm}

\def\eqref#1{equation~\ref{#1}}

\def\1{\bm{1}}

\DeclareMathAlphabet{\mathsfit}{\encodingdefault}{\sfdefault}{m}{sl}
\SetMathAlphabet{\mathsfit}{bold}{\encodingdefault}{\sfdefault}{bx}{n}

\def\gD{{\mathcal{D}}}

\usepackage{epsfig}
\usepackage{graphicx}

\usepackage{adjustbox}
\usepackage{array}
\usepackage{booktabs}
\usepackage{colortbl}
\usepackage{hhline}
\usepackage{multirow}
\usepackage{subcaption} %
\usepackage{floatflt}

\usepackage{amsmath,amsfonts,amssymb}
\usepackage{mathtools}  %
\usepackage{bm}
\usepackage{nicefrac}
\usepackage{microtype}

\usepackage{changepage}
\usepackage{extramarks}
\usepackage{fancyhdr}
\usepackage{lastpage}
\usepackage{soul}
\usepackage{xspace}

\usepackage{url}

\usepackage{enumerate}
\usepackage{todonotes} %
\usepackage{enumitem}  %

\usepackage{makecell}

\usepackage{pifont} %

\usepackage{algorithm,algpseudocode}

\usepackage{amsthm}
\usepackage{relsize}
\newcolumntype{L}[1]{>{\raggedright\let\newline\\\arraybackslash\hspace{0pt}}m{#1}}
\newcolumntype{C}[1]{>{\centering\let\newline\\\arraybackslash\hspace{0pt}}m{#1}}
\newcolumntype{R}[1]{>{\raggedleft\let\newline\\\arraybackslash\hspace{0pt}}m{#1}}

\newcommand{\sectapp}[1]{Appendix~\ref{#1}}

\newcommand{\eqn}[1]{Eq.~\ref{#1}}
\newcommand{\fig}[1]{Fig.~\ref{#1}}
\newcommand{\tbl}[1]{Table~\ref{#1}}

\newcommand{\ignore}[1]{}

\makeatletter
\DeclareRobustCommand\onedot{\futurelet\@let@token\@onedot}
\def\@onedot{\ifx\@let@token.\else.\null\fi\xspace}

\def\eg{e.g\onedot} 
\def\ie{i.e\onedot} 
 
\def\etc{etc\onedot}

\def\wrt{w.r.t\onedot}

\def\etal{et al\onedot}

\makeatother

\definecolor{MyDarkBlue}{rgb}{0,0.08,1}
\definecolor{MyDarkGreen}{rgb}{0.02,0.6,0.02}
\definecolor{MyDarkRed}{rgb}{0.8,0.02,0.02}
\definecolor{MyDarkOrange}{rgb}{0.40,0.2,0.02}
\definecolor{MyPurple}{RGB}{111,0,255}
\definecolor{MyRed}{rgb}{1.0,0.0,0.0}
\definecolor{MyGold}{rgb}{0.75,0.6,0.12}
\definecolor{MyDarkgray}{rgb}{0.66, 0.66, 0.66}

\newcommand{\model}{RSGs\xspace}

\newcommand{\mycell}[1]{\begin{tabular}[t]{@{}l@{}l}#1\end{tabular}}
\newcommand{\mycellc}[1]{\begin{tabular}[t]{@{}c@{}l}#1\end{tabular}}
\newcommand{\xhdr}[1]{{\noindent \bf #1}}

\newcommand{\modelplural}{RSGs\xspace}
\newcommand{\modelsingular}{RSG\xspace}

\newcommand{\astar}{\textit{A$^*$}\xspace}
\newcommand{\rrt}{\textsc{rrt}\xspace}
\newcommand{\fsm}{\textsc{fsm}\xspace}

\newcommand{\rationality}{\textrm{Rat}}
\newcommand{\states}{\mathcal{S}}
\newcommand{\actions}{\mathcal{A}}
\newcommand{\transition}{\mathcal{T}}
\newcommand{\cost}{\mathcal{C}}
\newcommand{\goal}{G}
\newcommand{\tasklang}{\mathcal{TL}\xspace}

\newcommand{\taskornospace}{\textit{\underline{or}}}
\newcommand{\taskthennospace}{\textit{\underline{then}}}
\newcommand{\taskand}{\textit{ \underline{and} }}
\newcommand{\taskor}{\textit{ \underline{or} }}
\newcommand{\taskthen}{\textit{ \underline{then} }}

\newcommand{\fsminit}{\textit{VI}}
\newcommand{\fsmterm}{\textit{VG}}

\usepackage[utf8]{inputenc} %
\usepackage{natbib}

\title{Learning Rational Subgoals from Demonstrations and Instructions}
\author{
    Zhezheng Luo\equalcontrib \textsuperscript{\rm 1},
    Jiayuan Mao\equalcontrib \textsuperscript{\rm 1},
    Jiajun Wu\textsuperscript{\rm 2},\\
    Tomás Lozano-Pérez\textsuperscript{\rm 1},
    Joshua B. Tenenbaum\textsuperscript{\rm 1},
    Leslie Pack Kaelbling\textsuperscript{\rm 1}
}
\affiliations{
    \textsuperscript{\rm 1} Massachusetts Institute of Technology \quad
    \textsuperscript{\rm 2} Stanford University\\
}

\usepackage{bibentry}

\begin{document}

\maketitle

\begin{abstract}
We present a framework for learning useful subgoals that support efficient long-term planning to achieve novel goals.  At the core of our framework is a collection of rational subgoals (\model), which are essentially binary classifiers over the environmental states. \model can be learned from weakly-annotated data, in the form of {\it unsegmented} demonstration trajectories, paired with abstract task descriptions, which are composed of terms initially unknown to the agent (\eg, \textit{collect-wood {\underline{then}} craft-boat {\underline{then}} go-across-river}). Our framework also discovers dependencies between \model, \eg, the task {\it collect-wood} is a helpful subgoal for the task {\it craft-boat}. Given a goal description, the learned subgoals and the derived dependencies facilitate off-the-shelf planning algorithms, such as \astar and \rrt, by setting helpful subgoals as waypoints to the planner, which significantly improves performance-time efficiency. Project page: \url{https://rsg.csail.mit.edu}
\end{abstract}
\section{Introduction}
\begin{figure*}[ht]
    \centering\small
    \includegraphics[width=0.9\textwidth]{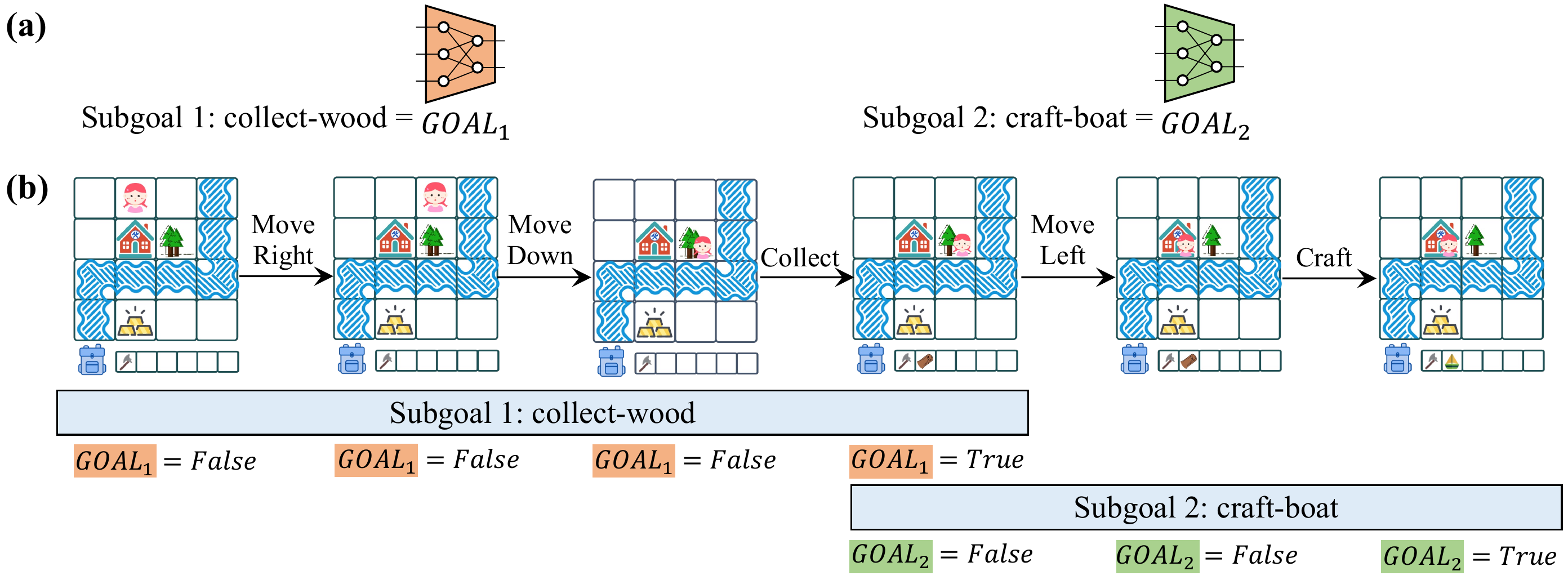}
    \caption{Interpreting a demonstration and its description in terms of \model: (a) Each \modelsingular is represented as a subgoal $\goal_o$. (b) The system infers a transition to the next subgoal if the $\goal$ condition is satisfied. Such transition rules can be used to interpret demonstrations and to plan for tasks that require multiple steps to achieve.
    }
    \label{fig:teaser}
\end{figure*} Being able to decompose complex tasks into subgoals is critical for efficient long-term planning. Consider the example in \fig{fig:teaser}: planning to craft a boat from scratch is hard, as it requires a long-term plan going from collecting materials to crafting boats, but it can be made easier if we know that {\it having an axe} and {\it having wood} are useful sub-goals. Planning hierarchically with these subgoals can substantially reduce the search required.
It is also helpful to understand the temporal dependencies between these subgoals, such as {\it having wood} being a useful subgoal to achieve {\em prior to} {\it crafting boat} makes long-term planning much more efficient.

In this work, we propose {\it Rational Subgoals} (\model), a framework for learning useful subgoals and their temporal dependencies from demonstrations. Our system learns with very weak supervision, in the form of a small number of {\it unsegmented} demonstrations of complex behaviors paired with abstract task descriptions.  The descriptions are composed of terms that are initially unknown to the agent, much as an adult might narrate the high-level steps when demonstrating a cooking recipe to a child. These action terms indicate important {\it subgoals} in the action sequence, and our agent learns to detect when these subgoals are true in the world, infer their temporal dependencies, and leverage them to plan efficiently.

Illustrated in \fig{fig:teaser}, our model learns from a dataset of paired but unaligned low-level state-action sequences and the corresponding abstract task description (\textit{collect-wood \underline{then} craft-boat \underline{then} go-across-river}).
For each action term $o$ (\eg, \textit{collect-wood}), our model learns a goal condition $\goal_o$, which maps any state to a binary random variable, indicating whether the state satisfies the goal condition. Given the training data, we decompose the observed trajectory into fragments, each of which corresponds to a ``rational'' sequence of actions for achieving a subgoal in the description.

While this model-based approach enables great generality in generating behaviors, it suffers from the slow online computation. To speed up online planning, we compute a dependency matrix whose entries encode which subgoals might be helpful to achieve before accomplishing another subgoal (\eg, {\it having wood} is a helpful subgoal for the task {\it crafting boat}, and thus the entry ({\it having wood}, {\it crafting boat}) will have a higher weight). During test time, given a final goal (\eg, {\it craft boat}) and the initial state, a hierarchical search algorithm is applied at both the subgoal level and the lower, environmental-action level.

The explicit learning of subgoals and their dependency structures brings two important advantages. First, the subgoal dependency allows us to explicitly set helpful subgoals as waypoints for planners. This significantly improves their runtime efficiency. Second, compared to alternative subgoal parameterizations such as reward functions, subgoals in the form of a state classifier allows us to use simple and efficient planners. For example, in continuous spaces, we can use Rapidly-exploring Random Trees (\rrt) to search for plans in the robot configuration space. These planers do not require training and generalize immediately to novel environments.

We evaluate \model in Crafting World~\citep{Chen2020Ask}, an image-based grid-world domain with a rich set of object crafting tasks, and Playroom~\citep{Konidaris2018Skills}, a 2D continuous domain with geometric constraints. Our evaluation shows that our model clearly outperforms baselines on planning tasks where the agent needs to generate trajectories to accomplish a given task. Another important application of \model is to create a language interface for human-robot communication, which includes robots interpreting human actions and humans instructing robots by specifying a sequence of subgoals. Our model enables compositional generalization through flexible re-composition of learned subgoals, which allows the robot to interpret and execute novel instructions.

\section{Rational Subgoal Learning and Planning}
We focus on learning rational subgoals from demonstration data and leveraging them for planning.
Formally, our training data is a collection of paired {\it unsegmented} demonstrations (\ie, state and action sequences) and abstract descriptions (\eg, \textit{collect-wood \underline{then} craft-boat}) composed of action terms (\textit{collect-wood}, \etc) and connectives (\taskthennospace, \taskornospace). Our ultimate goal is to recover the {\it grounding} (\ie, the corresponding subgoal specified by the action term) for each individual action term. These subgoals will be leveraged by planning algorithms to solve long-horizon planning problems.

We begin this section with basic definitions of the rational subgoal representations and the language $\tasklang$ for abstract descriptions. Second, we outline the planning algorithm we use to refine high-level instructions in $\tasklang$ into environmental actions that agents can execute, {\it given} the \modelplural. Although any search algorithms or Markov Decision Process (MDP) solvers are in principle applicable for our planning task, in this paper, we have focused on a simple extension to the A* algorithm. Next, we present the algorithm we use to {\it learn} \modelplural from data. Since we are working with unsegmented trajectories, the learning algorithm has two steps. It first computes a rationality score for individual actions in the trajectory based on the optimal plan derived from the A* algorithm. Then, it uses a dynamic programming algorithm to find the best segmentation of the trajectory and updates the parameters. Finally, we describe a dependency discovery algorithm for \modelplural and apply it to solve planning tasks given only a single goal action term (\eg, {\it collect-gold}), in contrast to the earlier case where there are detailed step-by-step instructions.

We call our representation {\it rational} subgoals because our learning algorithm is based on a {\it rationality} objective with respect to demonstration trajectories, and our planning algorithm chooses {\it rational} subgoals to accelerate the search.

Formally, a rational subgoal (\modelsingular) is a classifier that maps an environmental state $s$ to a Boolean value, indicating whether the goal condition is satisfied at $s$. Each \modelsingular has an atomic name $o$ (\eg, {\it collect-wood}), and the corresponding goal classifier is denoted by $\goal_o$. Depending on the representation of states, $\goal_o$ can take various forms of neural networks, such as convolutional neural networks (CNNs) for image-based state representations.

\begin{figure}[tp]
    \centering\small
    \includegraphics[width=0.45\textwidth]{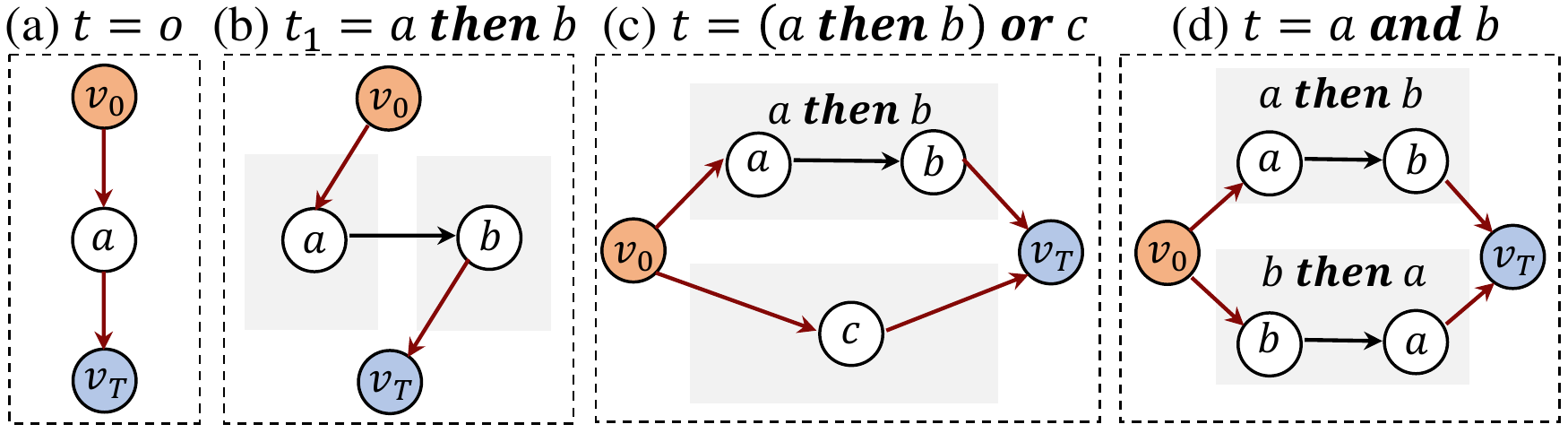}
    \caption{Illustrative example of how finite state machines (\fsm) are constructed from task descriptions. The super-starting node $v_0$ and the super-terminal node $v_T$ are highlighted.}
    \label{fig:statemachine}
\end{figure}

In both learning and planning, we will be using an abstract language to describe tasks, such as {\it collect-wood \underline{then} craft-boat}. These descriptions are written in a formal task language $\tasklang$. Syntactically, all atomic subgoals are in $\tasklang$;  and for all $t_1, t_2 \in \tasklang$, $(t_1 \taskthen{} t_2)$,  $(t_1 \taskor{} t_2)$,  and $(t_1 \taskand{} t_2)$ are in $\tasklang{}$. Semantically, a state sequence $\bar{s}$ satisfies a task description $t$, written $\bar{s} \models t$ when:
\begin{itemize}[noitemsep,topsep=0pt,parsep=0pt,partopsep=0pt,leftmargin=2em]
    \item If $t$ is a {\it \modelsingular} $o$, then the first state does not satisfy $\goal_o$, and the last state satisfies $\goal_o$. Note that this implies that the sequence $\bar{s}$ must have at least $2$ states.
    \item If $t = (t_1 \taskthen{} t_2)$ then $\exists 0 < j < n$ such that $(s_1, \ldots, s_j) \models  t_1$ and $(s_j, \ldots, s_n) \models  t_2$: task $t_1$ should be accomplished before $t_2$.
    \item If $t = (t_1 \taskor{} t_2)$ then $\bar{s} \models t_1$ or $\bar{s} \models t_2$: the agent should either complete $t_1$ or $t_2$.
    \item If $t = (t_1 \taskand{} t_2)$ then $\bar{s} \models (t_1 \taskthen{} t_2)$ or
    $\bar{s} \models (t_2 \taskthen{} t_1)$: the agent should complete both $t_1$ and $t_2$, but in any order ($t_1$ first or $t_2$ first)\footnote{The operator \taskand{} can be generalized be n-ary. In this case, accomplishing them in any order is considered accomplishing the composed task. For example, the task {\it mine-wood \taskand{} mine-gold \taskand{} mine-coal} allows the agent to accomplish all three subgoals in any order. Note that this is different from the specification with parenthesis: {\it (mine-wood and
mine-gold) and mine-coal}.}.
\end{itemize}
Note that the relation $\bar{s} \models t$ only specifies whether $\bar{s}$ completes $t$ but not how optimal $\bar{s}$ is. Later on, when we define the planning problem, we will introduce the trajectory cost.

Each task description $t \in \tasklang$ can be represented with a non-deterministic finite state machine (\fsm), representing the sequential and branching structures. Each $FSM_t$ is a tuple $(V_t,E_t,VI_t,VG_t)$ which are subgoal nodes, edges, set of possible starting nodes and set of terminal nodes. Each node corresponds to an action term in the description, and each edge corresponds to a possible transition of changing subgoals. \fig{fig:statemachine} illustrates the constructions for syntax in $\tasklang{}$, and we provide the follow algorithm for the construction.

\begin{itemize}
    \item \textbf{Single subgoal:} A single subgoal $s$ is corresponding \fsm with a single node i.e. $\fsminit_t=\fsmterm_t=V_t=\{s\}$, and $E_t=\emptyset$.
    \item $t_1~\taskthen{}~t_2$: We merge $\fsm_{t_1}$ and $\fsm_{t_2}$ by merging their subgoal nodes, edges and using $\fsminit_{t_1}$ as the new starting node set and $\fsmterm_{t_2}$ as the new terminal node set. Then, we add all edges from $\fsmterm_{t_1}$ to $\fsminit_{t_2}$. Formally,

    \begin{align*}
    &\fsm_{t_1 \taskthen{} t_2} = \\
    &(V_{t_1}\cup V_{t_2}, E_{t_1}\cup E_{t_2} \cup (\fsmterm_{t_1}\times \fsminit_{t_2}), \fsminit_{t_1}, \fsmterm_{t_2}),
    \end{align*}
    where $\times$ indicates the Cartesian product, meaning that each terminal node of $\fsm_{t_1}$ can transit to any starting node of $\fsm_{t_2}$.
    \item $t_1~\taskor{}~\cdots~\taskor{}~t_n$: Simply merge $n$ FSMs without adding any new edges. Formally,

    $$
    \fsm_{t_1~\textbf{or}~\cdots~\textbf{or}~t_n} = (\bigcup_i V_{t_i},\bigcup_i E_{t_i},\bigcup_i \fsminit_{t_i},\bigcup_i \fsmterm_{t_i})
    $$

    \item $t_1~\taskand{}~\cdots~\taskand{}~t_n$: Build $2^{n-1}n$ sub-FSMs over $n$ layers: the $i$-th layer contains $n\cdot \binom{n-1}{i-1}$ sub-FSMs each labeled by $(s,D)$ where $s$ is the current subgoal to complete (so this sub-FSM is a copy of $FSM_s$), and $D$ is the set of subgoals that have been previously completed. Then for a sub-FSM $(s_1,D_1)$ and a sub-FSM $(s_2,D_2)$ in the next layer, if $D_2=D_1\cup\{s_1\}$, we add all edges from terminal nodes of the first sub-FSM to starting nodes of the second sub-FSM. After building layers of sub-FSMs and connecting them, we set the starting nodes to be the union of starting nodes in the first layer and terminal nodes to be the union of terminal nodes in the last layer.
\end{itemize}

Note that our framework requires the starting and terminal nodes to be unique, but the construction above may output a \fsm with multiple starting/terminal nodes, so we introduce the virual super starting node $v_0$ and terminal node $v_T$ to unify them.

\paragraph{Remark.} In this paper, the language $\tasklang$ used for describing tasks covers LTL$_f$, a finite fragment of LTL that does not contain the {\it always} quantifier, so our fragment does not model task specifications that contain infinite loops. Finite LTL formulae can be converted to a finite automaton \citep{de2013linear}, represented using the \fsm.

\begin{figure}[tp]
    \centering
    \includegraphics[width=0.49\textwidth]{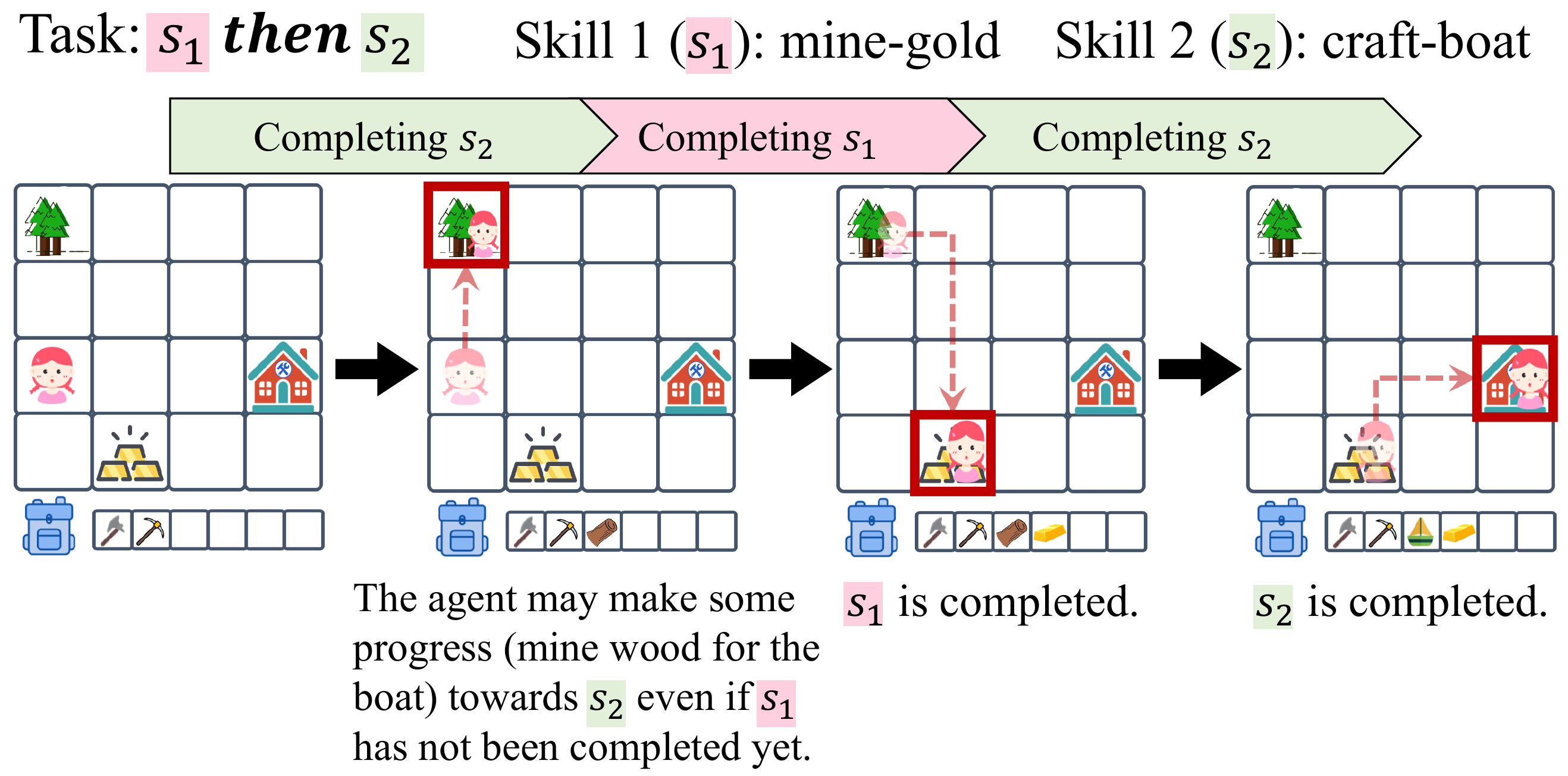}
    \caption{An example of optimal interleaving subgoals: $s_1$ is "mine gold", and $s_2$ is "craft boat". It is valid that the agent first goes to collect wood (for accompolishing $s_2$), and then mine gold (for accompolishing $s_1$), and finally crafts boat. In this case, the action sequences for completing $s_1$ and $s_2$ are interleaved. However, they can are be recognized as $s_2 \taskthen{} s_2$ because $s_1$ is accomplished before $s_2$.}
    \label{fig:skill-interleave}
\end{figure}
 \paragraph{Execution steps for different subgoals can interleave.}

\model does not simply run optimal policy for each individual subgoal sequentially. Rather, the semantic of $s_1 \taskthen{} s_2$ is: $s_1$ should be completed before $s_2$. It does not restrict the agent from making progress towards the subgoal  before the subgoal  is completed. In some case, such interleaving is necessary to obtain the globally optimal trajectory.

Consider the example shown in Figure \ref{fig:skill-interleave}, where $s_1$ is "mine-gold", and $s_2$ is "craft-boat". It is valid that the agent first goes to collect wood (for accompolishing $s_2$), and then mine gold (for accompolishing $s_1$), and finally crafts boat. In this case, the action sequences for completing $s_1$ and $s_2$ are interleaved. However, they can are be recognized as $s_1 \taskthen{} s_2$ because $s_1$ is accomplished before $s_2$.

\begin{figure*}[ht]
    \centering
    \includegraphics[width=0.8\textwidth]{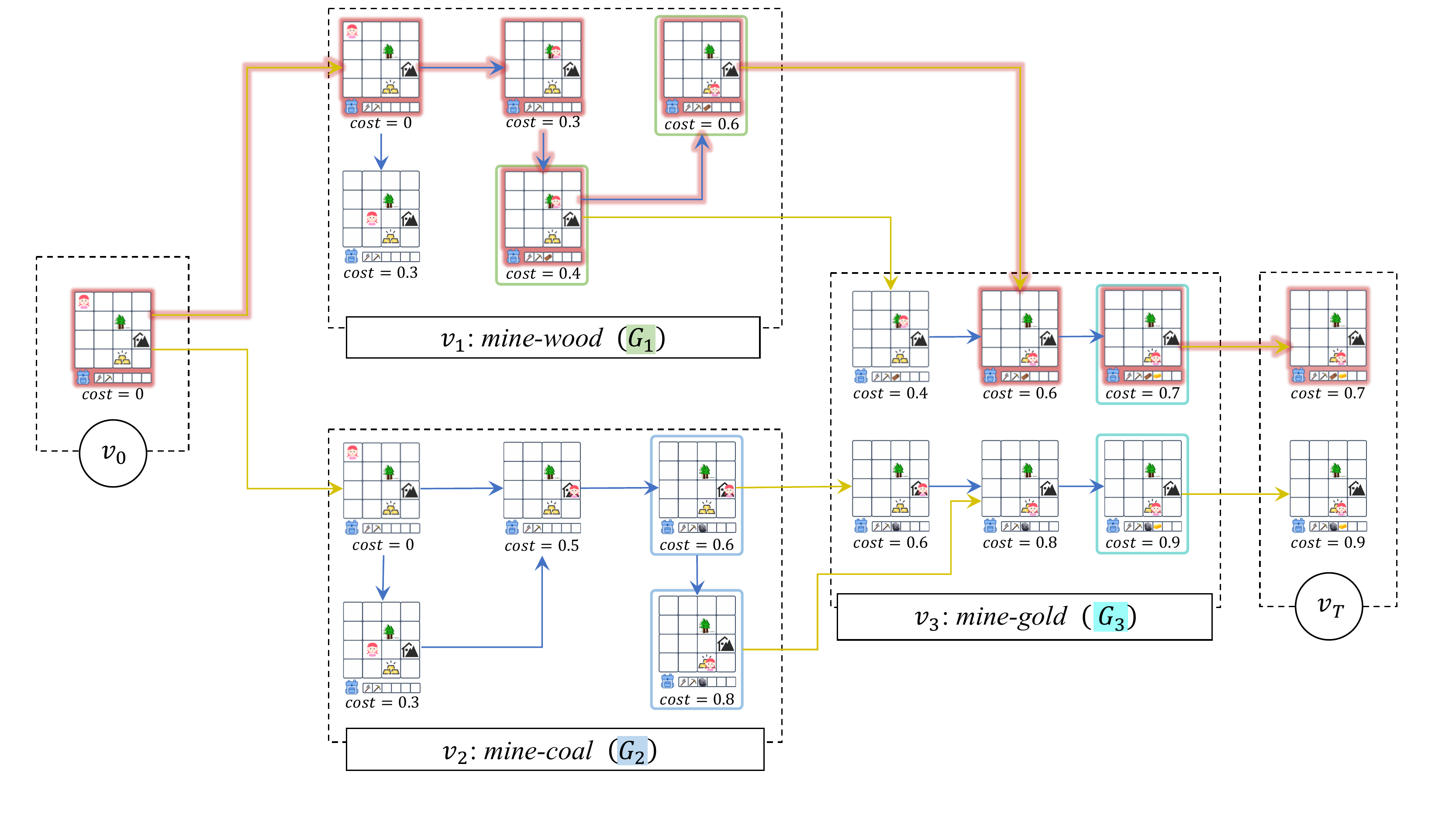}
    \caption{A running example of the \fsm-\astar algorithm for the task ``{\it (mine wood \taskor mine coal) \taskthen mine gold}.'' For simplicity, we only show a subset of states visited on each \fsm node. The blue arrows indicate transitions by primitive actions (in this example, each primitive action takes a cost of 0.1). The yellow arrows are transitions on the \fsm, which can only be performed when $\goal_v(\cdot)$ and $\goal_{v'}(\cdot)$ evaluates to False (in practice, the reward is computed as $-\left(\log \goal_v(\cdot) + \log \left(1-\goal_{v'}(\cdot)\right)\right)$). At the super-terminal node $v_T$, the state with minimum cost will be selected and we will back-trace the entire state-action sequence.}
    \label{fig:supp-astar-example}
\end{figure*}

\subsection{Planning with RSGs}
\label{ssec:planning}

We first consider the problem of planning an action sequence that satisfies a given task description $t$ written in $\tasklang$.  We assume that the external world is well modeled as a deterministic, fully observable decision process with a known state space, an action space, a transition function, and a cost function $\langle \states, \actions, \transition, \cost \rangle$ and that we have a set of goal classifiers $G_o$ parameterized by $\theta$.  Given a task $t$, we construct an \fsm representation and then compose it with the environment process to obtain an \fsm-augmented process $\langle \states_t, \actions_t, \transition_t, \cost_t\rangle$.  Concretely, $\states_t = \states \times V_t$, where $V_t$ is the set of nodes of \fsm constructed from task $t$. We then denote each task-augmented state as $(s, v)$, where $s$ is the environment state, and $v$ indicates the current subgoal. The actions $\actions_t = \actions \cup \fsm_t$, where each action either corresponds to a primitive action $a \in \actions$ or a transition in $\fsm_t$. An \fsm transition action indicates that the agent has achieved the current subgoal and will proceed to the next subgoal. We further define $\transition_t\left( (s, v), a \right) = (\transition(s, a), v)$ if $a$ is a primitive action in $\actions$, while $\transition_t\left( (s, v), a \right) = (s, v')$ if $a = (v, v') \in \fsm_t$ is an edge in the \fsm{}. The former are environmental actions. They only change the environmental state $s$ but do not change the current subgoal $v$. The latter, namely \fsm transitions, do not change the environmental state, but mark the current subgoal as completed and switch to the next one. Similarly, for the cost function,
\begin{center}
\begin{equation*}
\small
    \cost'\left( (s, v), a \right) = \begin{cases}
        \cost(s, a) & \text{if $a \in \actions$},\\
        -\lambda \left(\log \goal_v(s ;\theta) + \right. & \text{if $a = (v, v') \in \fsm_t$} \\
        ~~\left.\log \left(1 - \goal_{v'}(s ; \theta)\right)\right) &
    \end{cases}
\end{equation*}
\end{center}
where $\lambda$ is a hyperparameter. The key intuition behind the construction of $\cost_t$ is that the cumulative cost from $v_0$ to $v_T$ is the summation of all primitive action costs added to the log probability of the validity of subgoal transitions. At each subgoal transition, the state $s$ should satisfy the goal condition of the current \model but should not satisfy the goal condition of the next \model---which enforces the sequential constraints specified in the task.
In principle, when $\goal_v$ are Boolean-output classifiers, the cost is $0$ for a valid transition and $\infty$ for an invalid transition. In practice, we approximate the ``soft'' version of classifiers with neural networks: the outputs are in $[0,1]$, indicating how likely those conditions are to be satisfied.

Importantly, our formulation of the \modelsingular planning problem is different from planning for each individual action term and stitching the sub-plans sequentially. Concretely, we are finding a ``globally'' optimal plan instead of achieving individual subgoals in a locally optimal way. Thus, we allow complex behaviors such as making progress for a later subgoal to reduce the total cost. We include detailed examples in the supplementary material.

At the input-output level, our planner receives the a task description $t$ represented as an \fsm, an environmental transition model $\transition$, and a cost function $\cost$, together with a set of goal classifiers $\{\goal_o\}$ parameterized by $\theta$. It generates a sequence of actions $\bar{a}$ that is a path from $(s_0, v_0)$ to $(s_T, v_T)$ and minimizes the cumulative action costs defined by $\cost_t$. Here, $s_0$ is the initial environmental state, $v_0$ is the initial state of $\fsm_t$, $s_T$ is the last state of the trajectory, and $v_T$ is the terminal state of $\fsm_t$.

We make plans using slightly modified versions of \astar search, with a learned domain-dependent heuristic for previously seen tasks and a uniform heuristic for unseen tasks. This algorithm can be viewed as doing a forward search to construct a trajectory from a given state to a state that satisfies the goal condition. Our extension to the algorithms handles the hierarchical task structure of the \fsm.

Our modified \astar search maintains a priority queue of nodes to be expanded.
At each step, instead of always popping the
task-augmented state $(s, v)$ with the optimal evaluation, we first sample a subgoal $v$ uniformly in the \fsm{}, and then choose the priority-queue node with the smallest evaluation value among all states $(\cdot, v)$. This balances the time allocated to finding a successful trajectory for each subgoals in the task description.

Our hierarchical search algorithm also extends to continuous domains by integrating Rapidly-Exploring Random Trees (\rrt) \citep{lavalle1998rapidly}. We include the implementation details in the supplementary material.  Any state-action sequence produced by planning in the augmented model is legal according to the environment transition model and is guaranteed to satisfy the task specification $t$.

\paragraph{Example.}

\fig{fig:supp-astar-example} shows a running example of our \fsm-\astar planning given the task ``{\it mine wood \taskor mine coal \taskthen mine gold}'' from the state $s_0$ (shown as the left-most state in the figure).

\begin{enumerate}
    \item At the beginning, $(s_0,v_0)$ is expanded to the node $v_1$:{\it mine wood} and $v_2$:{\it mine coal} with \fsm transition actions at no cost.
    \item We expand the search tree node on $v_1$ and $v_2$ and compute the cost for reaching each states on $v_1$ and $v_2$.
    \item For states that satisfy the goal conditions for $v_1$ and $v_2$ (\ie, $\goal_1$ and $\goal_2$, respectively, and circled by green and blue boxes) and the initial condition for $v_3$ (\ie, $1-\goal_3$), we make a transition to $v_3$ at no cost (the states that do not satisfy the conditions can also be expanded to $v_3$ but with a large cost.
    \item Then search can be done in a similar way at $v_3$ and the states at $v_3$ that satisfy $\goal_3$ can reach $v_T$.
    \item For all states at $v_T$, we back-trace the state sequence with the minimum cost.
\end{enumerate}

\subsection{Learning \model from Unsegmented Trajectories and Descriptions}
\label{ssec:learning}

\begin{figure*}[t]
    \begin{minipage}{0.32\textwidth}
    \vspace{1em}
        \centering\small
        \includegraphics[width=0.99\textwidth]{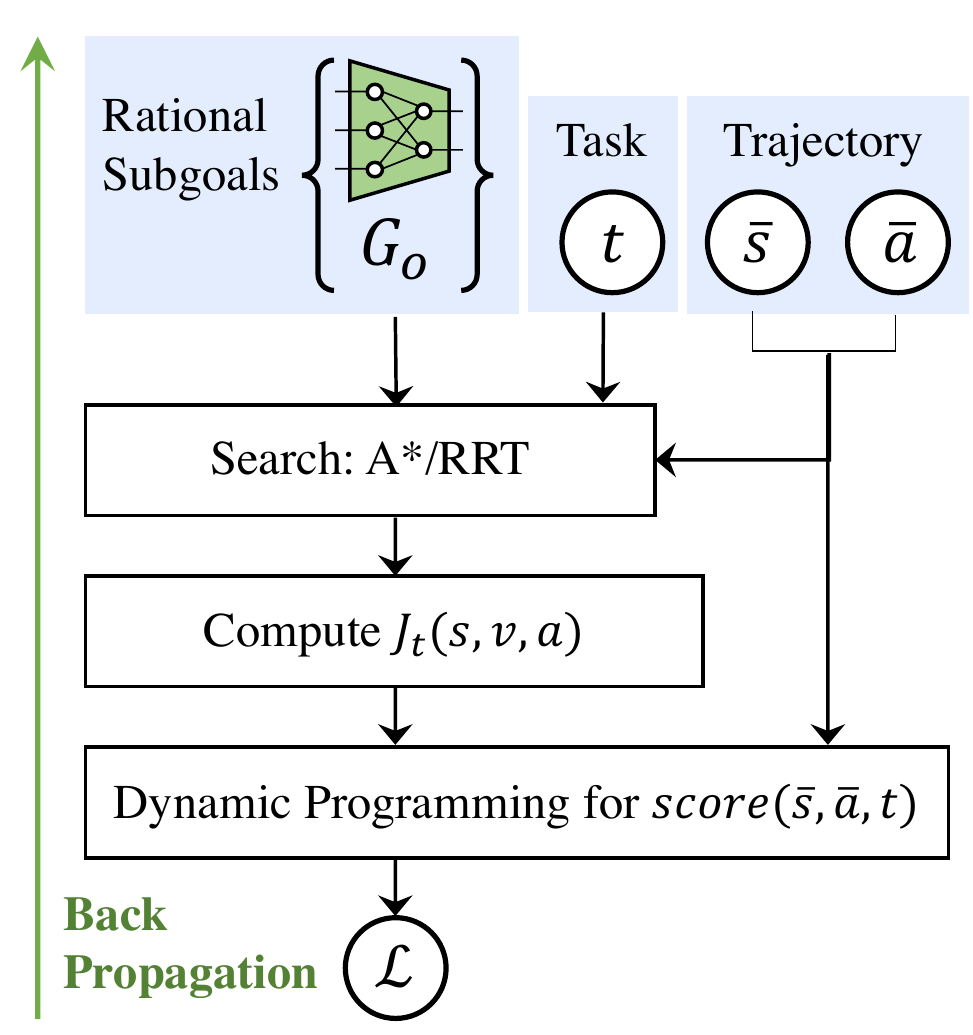}
        \caption{An overview of the training paradigm for \model. See text for details.}
        \label{fig:learning-diagram}
    \end{minipage}
    \hfill
    \begin{minipage}{0.67\textwidth}
    \begin{algorithm}[H]
        \centering
        \begin{algorithmic}
            \State Initiate the goal condition $G_o(\cdot;\theta)$
            \For{$(\bar{s},\bar{a}, t) \in D$}
            \For{$t'$ in candidate task descriptions}
                \State Apply A* search from all states in $\bar{s}$ with task $t'$ to compute a tree $T$.
                \For{each node $(s, v, a, t') \in T$ in reversed topological order}
                    \State Compute $J_{t'}(s,v,a;\theta)$ on the node using \eqn{eq:j-func}.
                \EndFor
                \For{each node $(s, v, a, t') \in T$ in reversed topological order}
                \State Compute $\rationality\left( s, v, a, t' ; \theta \right)$ for each tree node using \eqn{eq:rat}.
                \EndFor
                \State Compute $\textit{score}(\bar{s}, \bar{a}, t' ;\theta)$ using \eqn{eq:score} based on $\rationality$ values of nodes in $T$.
            \EndFor
            \State Compute the training objective $\mathcal{J}(\theta)$ using the score of all candidate task descriptions $t'$ using \eqn{eq:loss}.
            \State Update $\theta$ using gradient descent by maximizing $\mathcal{J}(\theta)$.
            \EndFor
        \end{algorithmic}
        \caption{Overview of the training paradigm in pseudocode.}
        \label{alg:training}
    \end{algorithm}
    \end{minipage}
\end{figure*}

We learn \modelplural from weakly-annotated demonstrations, in the form of {\it unsegmented} trajectories and paired task descriptions. The training dataset $\mathcal{D}$ contains tuples $(\bar{s}, \bar{a}, t)$ where  $\bar{s}$ is a sequence of environmental states, $\bar{a}$ is a sequence of actions, and $t \in \tasklang$ is a task description.

Our goal is to recover the grounding of subgoal terms from these demonstrations. At a high level, our learning objective is to find a set of parameters for the goal classifiers $\goal_o$ that {\it rationally} explain the demonstration data: the actions taken by the demonstrator should be ``close'' in some sense to the optimal actions that would be taken to achieve the goal. Let $\theta$ denote the collection of parameters in $\{\goal_o\}$. Thus, our training objective takes the following form:
\begin{align}
     \theta^* = \arg\max_\theta \frac{1}{|\mathcal{D}|} \sum_{(\bar{s}, \bar{a}, t) \in \mathcal D}  \textit{score}\left( \bar{s}, \bar{a}, t ;\theta\right) \;.
     \label{eq:theta}
\end{align}
The scoring function $\textit{score}$ combines the {\it rationality} of the observed trajectory with an additional term that emphasizes the appropriateness of {\sc fsm} transitions given $t$:
\begin{align}
    & \textit{score}(\bar{s}, \bar{a}, t ;\theta) \coloneqq \max_{\bar{v}}
    \mathlarger{\mathlarger{\{}}\log \prod_{i}\rationality\left( s_i, v_i, a_i, t ;\theta \right) + \nonumber \\
    & \sum_{\substack{(v_i, v_{i+1}) \in \\ \text{\fsm transitions}}} \left\{ \log \goal_{v_i}(s_i ; \theta) + \log \left(1 - G_{v_{i+1}}(s_i ; \theta)\right) \right\}\mathlarger{\mathlarger{\}}}
    \label{eq:score}
\end{align}
The rationality score measures the likelihood that the action $a \in \actions_t$ in state $(s, v)$ would have been chosen by a nearly-optimal agent, who is executing a policy that assigns a probability to an action based on the optimal cost-to-go for task $t$ in the {\sc fsm}-augmented model after taking it:
\begin{center}
\begin{align}
\rationality\left( s, v, a, t ; \theta \right) \coloneqq \frac{\exp \left( -\alpha \cdot J_t(s, v, a ; \theta) \right)}{\int_{x \in \mathcal{A}'} \exp \left( -\alpha \cdot J_t(s, v, x ; \theta) \right)},
\label{eq:rat}
\end{align}
\end{center}
where $\alpha$ is a hyperparameter called inverse rationality. The integral is a finite sum for discrete actions and can be approximated using Monte Carlo sampling for continuous actions.  If $\alpha$ is small, the assumption is that the demonstrations may be highly noisy; if large, then they are near optimal.

The cost-to-go (analogous to a value function) is defined recursively as
\begin{align}
J_t(s, v, a ; \theta) = \cost_t\left((s, v), a\right) + \max_{a' \in \actions_t} J_t\left(\transition'\left((s, v), a\right), a ;\theta\right).
\label{eq:j-func}
\end{align}
It need not be computed for the whole state space;  rather, it can be computed using the planner on a tree of relevant states, reachable from $(s_0, v_0)$.

Figure \ref{fig:learning-diagram} and Algorithm \ref{alg:training} summarize the learning process of \model. First, we perform a \astar search (or RRT for continuous domains) from the trajectory. Then, we backtrack in the search tree/RRT to compute the shortest distance from each node to the terminal state, $J_t$, so that $\text{Rat}(s_i,v_i,a_i,t;\theta)$ can be evaluated along the trajectory $\bar{s}, \bar{a}$.

At learning time, we can observe the environmental state and action sequence, but we cannot observe the {\sc fsm} states  or transitions.  To efficiently find the optimal \fsm states and transitions, given an environment state and action sequence as well as goal classifiers parameterized by the current $\theta$, we use a dynamic programming method.
Specifically, we will first label the FSM nodes from $0$ to $T$ by sorting them topologically. Next, we can use a two-dimensional dynamic programming with the transition equations based on $\text{Rat}$ and $G_v$ can find $\bar{v}$ that maximizes $score$. Concretely, let $f[i, j]$ denote the maximum score by aligning the trajectory $s_{i}, a_i, s_{i+1}, \cdots$ with the last $j$ nodes of the FSM. The dynamic programming algorithm iterates over $i$ in the reversed order. At each step, it tries to either assign the current $(s_i, a_i)$ pair to the current FSM node $j$, or to create a new transition from another FSM node $k$ to $j$. We present the detailed algorithm in the supplementary material.
Although the transition model we have discussed so far is deterministic, the methods can all be extended straightforwardly to the stochastic case, as also described in the supplement.

%
%
%
%
%
%
%
%
%
%
%
%
%
%
%
%
%

%

%

%

%

%
%
%
%
%
%
%

%

%

%

%
\iffalse
\begin{center}
\small
%
\begin{align*}
    \textit{score}(\bar{s}, \bar{a}, t) \coloneqq \max_{\bar{v}}
    \left\{  \log \prod_{i}\rationality\left( s_i, v_i, a'_i, t \right) + \sum_{\substack{(v_i, v_{i+1})~\text{is an \fsm transition}}} \left\{ \log \goal_{v_i}(s_i) + \log \left(1 - G_{v_i}(s_i)\right) \right\} \right\}
\end{align*}
%
\vspace{-1.5em}
\end{center}
\fi
%
%

To improve the optimization, we add a contrastive loss term, encoding the idea that, for each demonstration $(\bar{s}, \bar{a})$, the corresponding task description $t$ should have a higher rationality score compared to an unmatched task description $t'$, yielding the final objective to be maximized:
\begin{center}
\begin{align}
\small
\mathcal J(\theta) = & \sum_{(\bar{s}, \bar{a}, t) \in \mathcal D} \mathlarger{\mathlarger{\mathlarger{\mathlarger{(}}}} \textit{score}(\bar{s}, \bar{a}, t; \theta) \nonumber\\
& \quad + \left. \gamma \cdot \log \frac{\exp \left( \beta \cdot \textit{score}(\bar{s}, \bar{a}, t; \theta) \right)}{\sum_{t'} \exp \left( \beta \cdot \textit{score}(\bar{s}, \bar{a}, t'; \theta) \right)} \right),
\label{eq:loss}
\end{align}
\end{center}
where $t'$s are uniformly sampled negative tasks in $\tasklang$.
This loss function is fully differentiable \wrt $\theta$, which enables us to apply gradient descent for optimization. Essentially, we are back-propagating through two dynamic programming computation graphs: one that computes $J_t$ based on planning optimal trajectories given goal classifiers parameterized by $\theta$, and one that finds the optimal task-state transitions for the observed trajectory.

\begin{figure}[t]
    \centering\small
    \includegraphics[width=0.49\textwidth]{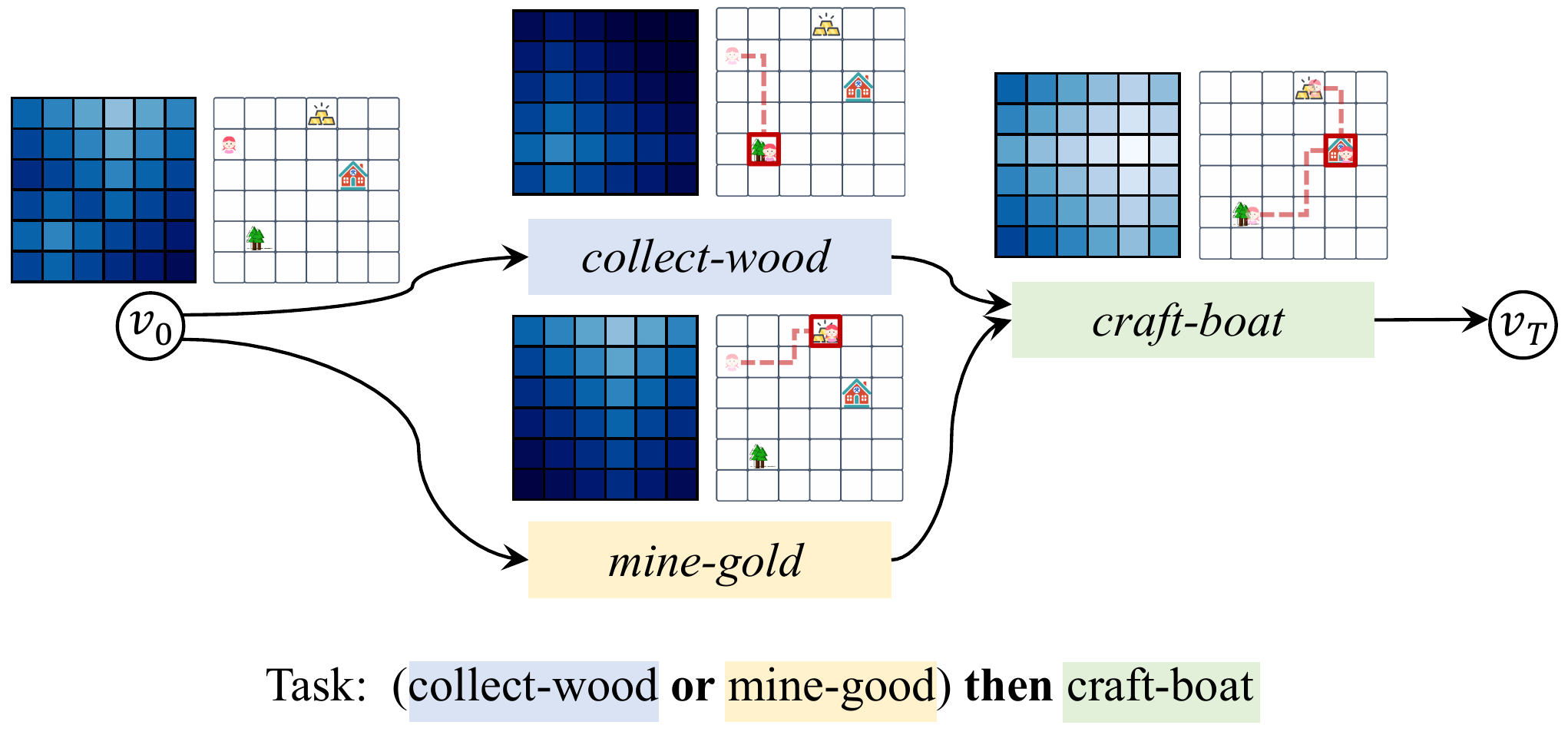}
    \caption{An example of the value function for task-augmented states on a simple \fsm. $\min_{a \in \actions} J_t\left(s, v, a\right)$ are plotted at each location at each \fsm node. Deeper color indicates larger cost. Red boxes and dotted lines illustrate the goal and a {\it rational} trajectory for each subgoal.}
    \label{fig:fsm-qvalues}
\end{figure}

\subsection{RSG Dependency Discovery and Planning}
\label{ssec:dependency}
Next, we describe our algorithm for planning with a single, final goal term (\eg, {\it craft-boat}) instead of step-by-step instructions. Since directly planning for the goal based on the corresponding goal classifier can be very slow due to the long horizon, our key idea here is to leverage the \model learned from data to perform a bilevel search. Our algorithm begins with discovering a dependency matrix between \model during training time. At performance time, we first use the discovered dependency model to suggest high-level plans, in the form of step-by-step instructions in $\tasklang$. Next, we use these instructions to plan for environmental actions using our planning algorithm.

For each possible subgoal $o$, we evaluate the associated learned goal classifier $G_o$ over all states along training trajectories that contain $o$. Next, we compute $\textit{first}(\bar{s}, o)$ as the smallest index $i$ such that $\goal_o(s_i)$ is true. If such $i$ does not exist (\ie, $\goal_o$ is never satisfied in $\bar{s}$) or $o$ is not mentioned in the task specification $t$ associated with $\bar{s}$, we define $\textit{first}(\bar{s}, o) = \infty$. For all tuples $(\bar{s}, o_1, o_2)$, we say $o_2$ is achieved {\it before} $o_1$ if neither $\textit{first}(\bar{s}, o_1)$ nor $\textit{first}(\bar{s}, o_1)$ is infinity, and $\textit{first}(\bar{s}, o_2) < \textit{first}(\bar{s}, o_1)$.

Let $\textit{bcount}(o_1, o_2)$ be the number of $\bar{s} \in \gD$ such that $o_2$ is achieved before $o_1$ in $\bar{s}$. We construct a dependency matrix $d$ by normalizing the $\textit{bcount}$ as:
\begin{align}
d(o_1, o_2) \triangleq \frac{\textit{bcount}(o_1, o_2)}{\sum_{o'} \textit{bcount}(o_1, o')},
\end{align}

where $o'$ sums over all \modelplural.

The derived dependency matrix can be interpreted as the probability that $o_2$ is a precondition for $o_1$. Now, recall that our task is to find an action sequence $\bar{a}$ that, starting from the initial state $s_0$, yields a new state $s_T$ that satisfies the given goal action term $g$, such as \textit{craft-boat}. Our high-level idea is to leverage the dependency matrix to suggest possible step-by-step instructions $t$, whose last action term is $g$. The planning algorithm will follow the suggested instructions to generate low-level plans $\bar{a}$.

Formally, we only consider instructions that are action terms connected by the {\it then} connective. Denote a candidate instruction $t = o_1 \taskthen o_2 \taskthen \cdots \taskthen o_k $. We define its priority as:
\begin{align}
\textit{priority}(t) = \lambda ^ k \prod_{i=1}^{k-1} \left(1-\prod_{j=i+1}^k \left(1-d(o_j, o_i)\right)\right),
\label{eq:priority}
\end{align}
where $\lambda$ is a length bias constant which is set to $0.9$ because we prefer shorter instructions.

Given the candidate instructions, we run the planning algorithm for these instructions. We prioritize instructions $t$ with high priorities $\textit{priority}(t)$, and these instructions are generated by a search approach (Algorithm \ref{alg:ins-gen}) from the given final goal. The limit of instruction length, $\textit{length\_limit}$, is set to $6$ for our experiment.. For more complicated domains, a promising future direction is to learn a full abstract planning model (symbolic or continuous) based on the subgoal terms learned from demonstrations.

\begin{algorithm}
\begin{algorithmic}
    \State Build a priority queue of instructions $H$.
    \State $H \gets \{\textit{final\_goal}\}$
    \While{$H$ is not empty}
        \State $t \gets H.pop()$
        \State Run A* search on task $t$.
        \If{the A* search finds a solution}
        \State \textbf{Return} the solution.
        \EndIf
        \If{$\textit{length}(t) \leq \textit{length\_limit}$}
            \For{$o \in O$}
                \If{$o \notin t ~\textbf{and}~ \exists o' \in t. d(o',o) > 0$}
                    \State $H.push(o \taskthen t)$~~\# See  \eqn{eq:priority}.
                \EndIf
            \EndFor
        \EndIf
    \EndWhile
\end{algorithmic}

\caption{Overview of the search algorithm given only the final goal.}
\label{alg:ins-gen}
\end{algorithm}

\newcommand{\crafting}{Crafting World\xspace}
\newcommand{\bcfsm}{BC-\fsm}

\section{Experiments}

We compare our model with other subgoal-learning approaches in  \crafting~\citep{Chen2020Ask}, a Minecraft-inspired crafting environment, and Playroom~\citep{Konidaris2018Skills}, a 2D continuous domain with geometric constraints.

\paragraph{\crafting.} In \crafting, the agent can move in a 2D grid world and interact with objects next to it, including picking up tools, mining resources, and crafting items. Mining in the environment typically requires tools, while crafting tools and other objects have their own preconditions, such as being close to a workstation or holding another specific tool. Thus, crafting a single item often takes multiple subgoal steps. There are also obstacles such as rivers (which require boats to go across) and doors (which require specific keys to open).

\begin{figure}[ht]
    \begin{minipage}{0.24\textwidth}
    \centering
    \includegraphics[width=0.8\textwidth]{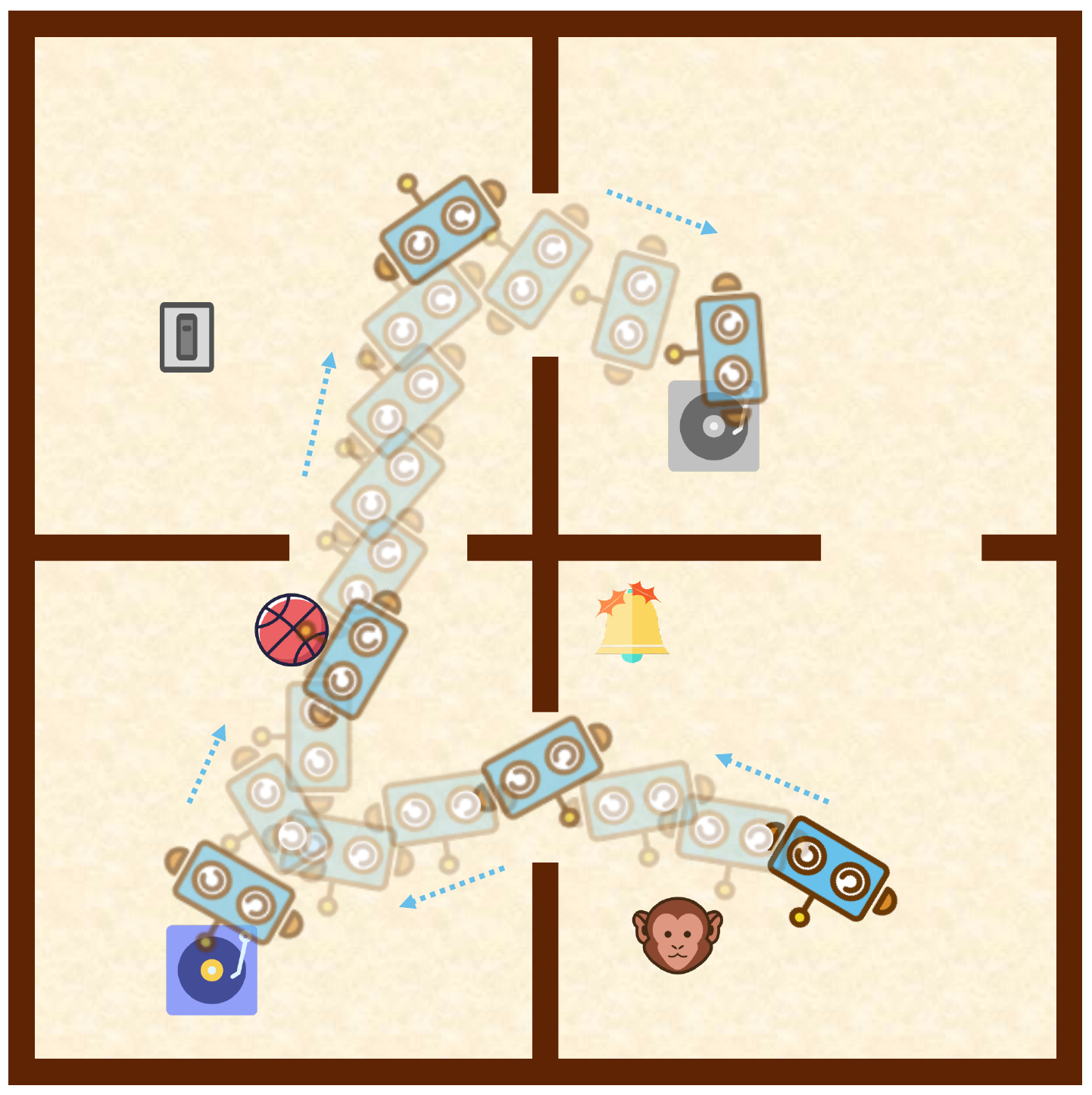}
    \end{minipage}
    \hfill
    \begin{minipage}{0.21\textwidth}
    \caption{An illustration of the Playroom environment and a trajectory for the task: {\it turn-on-music \underline{then} play-with-ball \underline{then} turn-off-music}.}
    \label{fig:playroom}
    \end{minipage}
\end{figure}
We define 26 primitive tasks, instantiated from templates of {\it grab-X}, {\it toggle-switch}, {\it mine-X}, and {\it craft-X}. While generating trajectories, all required items have been placed in the agent's inventory. For example, before mining wood, an axe must be already in the inventory. In this case, the agent is expected to move to a tree and execute the mining action. We also define 26 {\it compositional} tasks composed of the aforementioned primitive tasks. For each task, we have 400 expert demonstrations.

All models are trained using tuples of task description $t$ and expert state-action sequences $(\bar{s}, \bar{a})$. In particular, we train all models on primitive and {\it compositional} tasks and test them on two splits: {\it compositional} and {\it novel}. The {\it compositional} split contains novel state-action sequences of previously-seen tasks. The novel split contains 12 novel tasks, where primitive tasks are composed in ways never seen during training (\ie, not in the 26 tasks from the {\it compositional} split).

\paragraph{Playroom.} Our second environment is Playroom~\citep{Konidaris2018Skills}, a 2D maze with continuous coordinates and geometric constraints. \fig{fig:playroom} shows an illustrative example of the environment.
Specifically, a 2D robot can make moves in a small room with obstacles. The agent has three degrees of freedom (DoFs): $x$ and $y$ direction movement, and a 1D rotation.
The environment invalidates movements that cause collisions between the agent and the obstacles. Additionally, there are six objects randomly placed in the room, which the robot can interact with. For simplicity, when the agent is close to an object, the corresponding robot-object interaction will be automatically triggered.

Similar to the \crafting, we have defined six primitive tasks (corresponding to the interaction with six objects in the environment) and eight compositional tasks (\eg, {\it turn-on-music \underline{then} play-with-ball}). We have designed another eight novel tasks, and for each task, we have 400 expert demonstrations. We train different models on rational demonstrations for both the primitive and compositional tasks, and evaluate them on the compositional and novel splits.

\subsection{Baselines}

We compare our \model, which learns goal-based representations, with two baselines using different underlying representations: IRL methods learn reward-based representations, and behavior cloning methods directly learn policies. The implementation details are in the supplementary material.

Our max-entropy inverse reinforcement learning~\citep[IRL;][]{ziebart2008maximum} baseline learns a task-conditioned reward function by trying to explain the demonstration. For planning, we use the built-in deep-Q-learning algorithm. The behavior cloning~\citep[BC;][]{torabi2018behavioral} baseline directly learns a task-conditioned policy that maps the current state and the given task to an environment primitive action. \bcfsm is the BC algorithm augmented with our \fsm{} description of tasks. Compared with \model, instead of segmenting the demonstration sequence based on rationality, \bcfsm segments them based on how consistent each fragment is with the policy for the corresponding action term.

\subsection{Results}

To evaluate planning, each algorithm is given a new task $t$, either specified in  $\tasklang$, or as a black-box goal state classifier, and generates a trajectory of actions to complete the task.

\xhdr{Planning with instructions.} \tbl{tab:result:planning} summarizes the results. Overall, \model outperforms all baselines. On the {\it compositional} split, our model achieves a nearly perfect success rate in the \crafting ($99.6\%$). Comparatively, although the tasks have been presented during training of all baselines, their scores remain below $40\%$.

\begin{table}[tp]
    \centering
    \small
    \begin{tabular}{l l c cc cc}
        \toprule
       \multirow{2}{*}{Model} & \multirow{2}{*}{\mycell{Task\\Input}} & \multirow{2}{*}{\mycell{Env.\\Tran.}} &
        \multicolumn{2}{c}{\crafting} & \multicolumn{2}{c}{Playroom} \\
        \cmidrule{4-5} \cmidrule{6-7}
        & & & Com. & Novel & Com. & Novel \\
       \midrule
       IRL & Lang. & Y & 36.5 & 1.8 & 28.3 & 9.6  \\
       BC  & Lang. & N & 11.2 & 0.8 & 15.8 & 4.8 \\
       \bcfsm & FSM & N & 5.2 & 0.3 & 38.2 & 31.5 \\
       \midrule
       \model & FSM & Y & \bf 99.6 & \bf 97.8 & \bf 82.0 & \bf 78.2 \\
       \bottomrule
    \end{tabular}
    \captionof{table}{Results of the planning task, evaluated as the success rate of task completion. IRL and BC take raw task specification and process them with LSTM, while \bcfsm and \model uses the FSM directly. \model and IRL use the environmental transition model during training while BC and \bcfsm dot not. The maximum number of expanded nodes for all planners is 5,000. All models are trained on the {\it compositional} split, and tested on the {\it compositional} and the {\it novel} split.}
    \label{tab:result:planning}
\end{table}
\begin{figure*}[ht]
    \centering
    \includegraphics[width=0.99\textwidth]{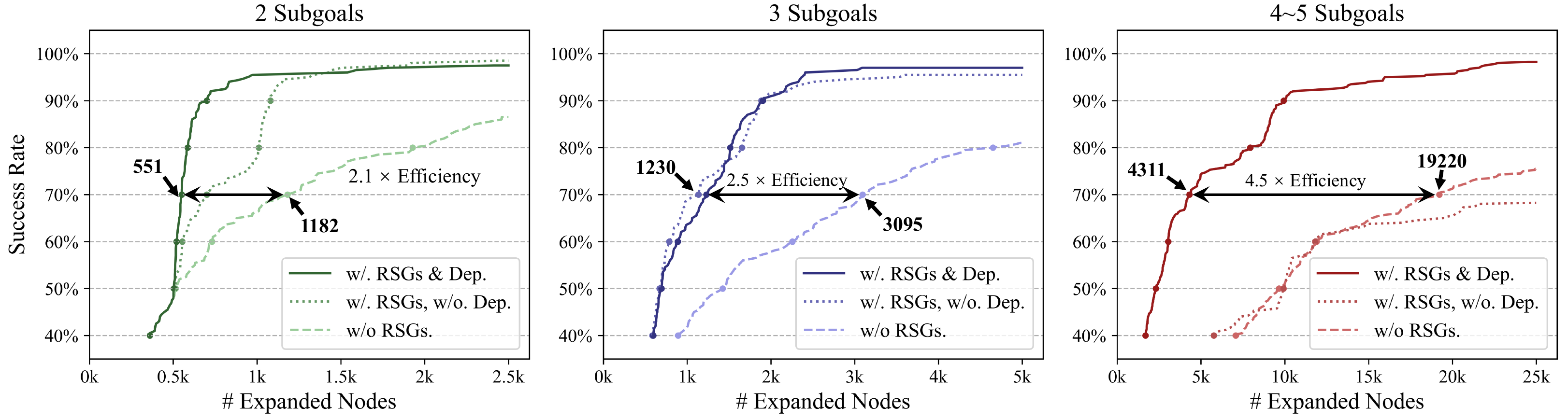}
    \caption{\model applied to planning with a final goal. We do evaluation on 3 groups of planning tasks in the \crafting environment. We use 100 random initial states for each task. Each search method can expand up to 25,000 nodes.}
    \label{fig:plan-search}
\end{figure*}
On the {\it novel} split, \model outperforms all baselines by a larger margin than on the {\it compositional} split. We observe that since {\it novel} tasks contain longer descriptions than those in the {\it compositional} set, all baselines have a success rate of almost zero.
Compared with IRL methods, the more compositional structure in our goal-centric representation allows it to perform better.
Meanwhile, a key difference between behavior cloning methods (BC and \bcfsm) and ours is that BC directly applies a learned policy, while our model runs an A* search based on the learned goal classifier and leverages the access to the transition model. This suggests that learning goals is more sample-efficient than learning policies in such domains and generalizes better to new maps.

Our model can be easily applied to environments with image-based states, simply by changing the inputs of $I_o$ and $G_o$ models to images. We evaluate our model in an image-based \crafting environment. It achieves 82.0\% and 78.2\% success rates on the compositional and novel splits, respectively. Comparatively, the best baseline \bcfsm gets 38.2\% and 31.5\%. Details are in the supplementary material.

\paragraph{Planning with goals.}
We also evaluate \model on planning with a single goal action term.
These problems require a long solution sequence, making them too difficult to solve with a blind search from an initial state.
Since there is no task specification given, in order to solve the problems efficiently, it is critical to use other dependent \model for search guidance. We use 8 manually designed goal tests, each of which can be decomposed into 2--5 subgoals. We run our hierarchical search based on \model and the discovered dependencies.

We compare this method with two baselines: a blind forward-search algorithm, and a hierarchical search based on \model without discovered dependencies (\ie, by setting the dependency matrix as a uniform distribution). We test all three methods on 100 random initial states for each task. \fig{fig:plan-search} summarizes the result.
Overall, \model with discovered dependencies enables efficient searches for plans. On easier tasks (2 or 3 subgoals), search with \model and dependencies has a similar runtime as the baseline that searches without dependencies. Both of them outperform the blind-search baseline (about 2.4$\times$ more efficient when reaching a 70\% success rate).
However, when the task becomes complex (4 or 5 subgoals), searching with \model and the discovered dependencies significantly outperforms other alternatives. For example, to reach a 70\% success rate, searching with \model needs only 4,311 expanded nodes. By contrast, searching without \model needs 19,220 (4.5$\times$) nodes. Interestingly, searching with \model but without discovered dependencies performs worse than the blind-search baseline. We hypothesize that this is because it wastes time on planning for unreasonable instructions. Overall, the effectiveness of \model with discovered dependencies grows as the complexity of tasks grows.
\section{Related Work}

\xhdr{Modular policy learning and planning.}
Researchers have been learning modular ``policies'' by simultaneously looking at trajectories and reading task specifications in the form of action term sequences~\citep{corona-etal-2021-modular,andreas2017modular,andreas-klein-2015-alignment}, programs ~\citep{sun2020program}, and linear temporal logic (LTL) formulas~\citep{bradley2021learning,icarte-teaching-2018,tellex2011understanding}. However, they either require additional annotation for segmenting the sequence and associating fragments with labels in the task description~\citep{corona-etal-2021-modular,sun2020program}, or cannot learn models for planning~\citep{tellex2011understanding}.
By contrast, \model learns useful subgoals from demonstrations. We use a small but expressive subset of LTL for task description, and jointly learn useful subgoals and segment the demonstration sequence.

Our subgoal representation is also related to other models in domain control knowledge~\citep{de2011learning}, goal-centric policy primitives~\citep{park2020inferring}, macro learning~\citep{newton2007learning}, options and hierarchical reinforcement learning~\citep[HRL;][]{sutton1999between,dietterich2000hierarchical,barto2003recent,mehta2011hierarchical}, and methods that combine reinforcement learning and planning~\citep{segovia2016planning,winder2020planning}. However, the execution of subgoals in \model is fundamentally different from options: each option has a policy that we can follow to achieve the short-term goal, while subgoals in \model should be refined with segments of primitives by planning algorithms. Our planning algorithm is similar to other approaches: \cite{de2011learning,botvinick2014model,winder2020planning}, but they do not leverage discovered dependencies between subgoals.

\xhdr{Learning from demonstration.} %
Learning from demonstration generally refers to building agents that can interact with the environment by observing expert demonstrations (e.g., state-action sequences). Techniques for learning from demonstration can be roughly categorized into four groups: policy function learning~\citep{chernova2007confidence,torabi2018behavioral}, cost and reward function learning~\citep{wulfmeier2015,ziebart2008maximum}, generative adversarial learning~\citep{ho2016generative,liu2022plan}, and learning high-level plans~\citep{ekvall2008robot,konidaris2012robot}. We refer to \citet{argall2009survey} and \citet{ravichandar2020recent} as comprehensive surveys. In this paper, we learn useful subgoals that support planning, and compare our model with methods that directly learn policies and cost functions.
Moreover, unlike those who use similarities between different actions~\citep{niekum2012learning} to segment demonstrations, in \model, we segment the demonstration with associate action terms by rationality assumptions of the agent.

\xhdr{Inverse planning.} Our model is also related to inverse planning algorithms that infer agent intentions from behavior by finding a task description $t$ that maximizes the consistency between the agent's behavior and the synthesized plan~\citep{baker2009action}. While existing work has largely focused on modeling the rationality of agents~\citep{baker2009action,zhi2020online} and more expressive task descriptions~\citep{Shah2018bayesian}, our focus is on leveraging the learned subgoals and their dependencies to facilitate agent planning for novel tasks.

\xhdr{Unsupervised subgoal discovery.} Our method is also related to approaches for discovering subgoals from unlabelled trajectories~\citep{paul2019learning, tang2018subgoal,kipf2019compile,lu2021learning,gopalakrishnan2021unsupervised}, mostly based on the assumption that the trajectory can be decomposed into segments, and each segment corresponds to a subgoal. Some other approaches for discovering subgoals are to detect ``bottleneck'' states ~\citep{menache2002q,csimcsek2005identifying} based on the state transition graphs. RSG differs from these works in that we focus on learning the grounding of action terms defined in task descriptions. Thus, RSGs are associated with action terms and thus can be recomposed by human users to describe novel tasks. It is a meaningful future direction to combine learning from trajectory-only data and trajectories with descriptions to improve the data efficiency.

\section{Conclusion}

We have presented a subgoal learning framework for long-horizon planning tasks. The rational subgoals (\model) can be learned by observing expert demonstrations and reading task specifications described in a simple task language~$\tasklang$. Our learning algorithm simultaneously segments the trajectory into fragments corresponding to individual subgoals, and learns planning-compatible models for each subgoal. Our experiments suggest that our framework has strong compositional generalization to novel tasks.

\paragraph{Limitation.}
The assumption of a deterministic environment has allowed us to focus on the novel RSG formulation of subgoal models. For domains with substantial stochasticity, the high-level concepts of RSGs could be retained (\eg, rationality), and algorithmic changes may be required such as replacing maximum entropy IRL with maximum causal entropy~\citep{ziebart2010modeling}. Another limitation of \model is that it can not leverage trajectories without labeled task descriptions. Future work may consider the jointly learning of subgoals and subgoal structures of tasks~\citep{vazquez2018learning,chou2022learning}.

\vspace{1em}
\xhdr{Acknowledgement.}
We thank Yunyun Wang for giving advice on making figures. We thank all group members of the MIT Learning \& Intelligent Systems Group for helpful comments on an early version of the project. This work is in part supported by NSF grant 2214177, AFOSR grant FA9550-22-1-0249, ONR MURI grant N00014-22-1-2740, the MIT-IBM Watson Lab, the MIT Quest for Intelligence, the Center for Brain, Minds, and Machines (CBMM, funded by NSF STC award CCF-1231216), the Stanford Institute for Human-Centered Artificial Intelligence (HAI), and Analog, Amazon, JPMC, Meta, Salesforce, and Samsung. Any opinions, findings, and conclusions or recommendations expressed in this material are those of the authors and do not necessarily reflect the views of our sponsors.

{
\bibliography{reference}
}

\newpage

\onecolumn

\begin{center}
\bf \LARGE Supplementary Material for \\
Learning Rational Subgoals from Demonstrations and Instructions
\end{center}

\vspace{2em}

\appendix
\newcommand{\playroom}{Playroom\xspace}
\newcommand{\dand}{\textit{\underline{and}}~}
\newcommand{\dor}{\textit{\underline{or}}~}
\newcommand{\dthen}{\textit{\underline{then}}~}

First, we elaborate how \astar search is performed on the \fsm-augmented transition models. We also discuss representation choices of \model, as well as the optimality, complexity and scalability of the search algorithms. Recall that we are using dynamic programming obtain deterministic transitions, and there are other formulations such as using stochastic transitions, we talk about the comparison between our formulation and others in this section. In addition, we provide details for the dependency discovering algorithm and hierarchical search algorithm used in planning for final goals.

Second, we discuss details about datasets and how we process the data, including how the features are extracted from the state representations. We also provide the list of task descriptions covered in each data split.

Next, we provide implementation details for baselines, and then discusses the limitation and future work of \model.

\section{Implementation Details of \model}\label{app:ratskill-details}

\paragraph{Re-parameterize $1-G_o(\cdot)$ using a separate neural network.}
In practice, instead of directly using $1-G_o(\cdot)$ to evaluate the probability that a subgoal has not been met, we parameterize $1-G_o(\cdot)$ using a separate neural network $I_o(\cdot)$ that has the same architecture as $G_o(\cdot)$. We observe that this re-parameterization stabilizes the training. In performance time, we will only be using the goal classifier $G_o$ and ignores the $I_o$.

Empirically, we find that when using a single subgoal classifier instead of separate classifiers for $I$ and $G$, some classifiers usually get stuck at local optima. As a result, the overall planning performance, on average, drops from 99.6\% to 75\%. We hypothesize that this is because using separate parameterization allows a broader set of possible solutions to the original problem, which are practically equivalently helpful for planning. As a concrete example, consider the subgoal "mining-wood."

If we use only the $G$ and $1-G$ parameterization, the only feasible solution is:

\[
G_1 = [\text{wood in inventory}]
\]

However, if we use separate parameterizations, the following solution will also be accepted:

\[
I_2 = [\text{tree on map and axe in inventory and wood not in inventory}]
\]
\[
G_2 = [\text{tree on map and axe in inventory and wood in inventory}]
\]

Note that, during planning, both classifiers $G_1$ and $G_2$ have the same effects, but the relaxed parameterization allows a broader set of solutions.

\subsection{\fsm-\astar}
\label{app:ratskill-details:astar}

We have implemented a extended version of the \astar algorithm to handle \fsm states in \crafting.

\paragraph{\astar at each \fsm node.}
We start with the \astar search process happening at each \fsm node. For a given \fsm state, the \astar search extends the tree search in two stages. The first stage lasts for $b=3$ layers during training and $b=4$ layers during testing. In the first $b$ layers of the search tree, we run a Breadth-First-Search so that every possible path with length $b$ is explored. Then on the second stage lasts for $c=15$ layers during training and $25$ layers in testing. In layer $d \in [b+1, b+c]$, we run A* from the leaves in the first stage based on the heuristic for each node. By enforcing the exploration at the early stage, we avoid imperfect heuristic from misguiding the A* search at the beginning.
For each \fsm node $v$ and each layer $d$, we only keep the top $k=10$. Finally, we run the value iteration on the entire search tree.

To accelerate this search process, for all tasks $t$ in the training set, we have initialized a dedicated value approximator $V_t(\bar{s})$, conditioned on the historical state sequence. During training, we use the value iteration result on the generated search tree to supervise the learning of this approximator $V_t$. Meanwhile, we use the value prediction of $V_t$ as the heuristic function for node pruning. During test, since we may encounter unseen tasks, the \astar-\fsm search uses a uniform heuristic function $h\equiv 0$.

\paragraph{Search on an \fsm.}
For a given initial state $s_0$ and task description $t$, we first build $\fsm_t$ and add the search tree node $(s_0,v_0)$ to the search tree, where $v_0$ is the initial node of the \fsm. Then we expand the search tree nodes $(s, v)$ by a topological order of $v$. It has two stages. First, for each \fsm node $v$, we run up to 5000 steps of \astar search. Next, for all search tree nodes $(s, v)$ at \fsm node $v$, we try to make a transition from $(s, v)$ to $(s, v')$ where $(v, v')$ is a valid edge in $\fsm_t$. Finally, we output a trajectory ending at the \fsm node $v_T$ with minimum cost.

\paragraph{Optimality of the A* algorithm on \fsm.}
In the current implementation, \model might return sub-optimal solutions even with a perfect heuristic, because \model balance the expanded nodes across all FSM nodes: it first samples an FSM node and then expand a search tree node with the best heuristic value on that node.

The optimality can be guaranteed by either of the following simple modifications, although at the cost of possibly increasing the running time:

\begin{itemize}
\item Always expand the search node with the globally best admissible heuristic value. (Because our heuristic is learned, this may not be practical.)
\item Keep expanding nodes, even after finding a plan, until none of the unexpanded search tree nodes across all subgoal nodes in the FSM have better heuristic values than the current best solution.
\end{itemize}

\subsection{Transitions on \fsm}
\label{app:ratskill-details:transitions}

\begin{algorithm}
\begin{algorithmic}
    \State Initialize $f[1..n, 0..T]$ to $-\infty$
    \State Topological sort all FSM nodes $v_{0..T}$ so that for all $0 \leq i < j\leq T$, there is no path from $v_j$ to $v_i$ on FSM. Clearly $v_0$ is still the start node and $v_T$ is the terminate node.
    \State $f[n,T] \leftarrow 0$
    \For{$i = n..1$}
        \For{$j = T..0$}
            \If{$i < n$}
                \State $f[i,j] \leftarrow \rationality(s_i,v_j,a_i,t;\theta) + f[i+1,j]$
            \EndIf
            \For $k = 0..T$:
                \If{$(v_j,v_k) \in$ FSM transitions}:
                    \State $f[i,j]\leftarrow \max${$f[i,j], \log G_{v_j}(s_i;\theta)+\log (1-G_{v_k}(s_i;\theta) + f[i+1,k]$}
                \EndIf
            \EndFor
        \EndFor
    \EndFor
    \State Return $score(\bar{s}, \bar{a}, t;\theta)=f[1,0]$
\end{algorithmic}

\caption{The dynamic programming for computing $score(\bar{s}, \bar{a}, t;\theta)$ given $G_o(\cdot;\theta)$ and $\rationality(s_i,v_i,a_i,t;\theta)$.}
\label{alg:dp}
\end{algorithm}

When encoding transitions on \fsm, we use dynamic programming \footnote{The dynamic programming is similar to Dynamic time warping(DTW) warping trajectories into sequential subgoals, but the “cost” is computed at each segment instead of at matched positions.} to select a transition that maximize our $score$. Algorithm \ref{alg:dp} shows the pseudo-code of the dynamic programming. If we consider $I_v$ and $G_v$ as ``soft'' probabilities, the computation of $score$ finds $\bar{s}'$ and $\bar{a}'$ that maximize the rationality of primitive actions and the likelihood that FSM transitions are successful. We use $score(\bar{s}, \bar{a}, t)$ to rank all candidate tasks.

There is another formulation which is to consider the stochastic transitions using $G_v$ as probabilities. These two formulations can be unified using a framework of latent transition models, though they are computed using different DP algorithms and may lead to different results.

First of all, these two formulations are equivalent when the goal classifiers are binary (0/1). When the classifiers are approximated by "soft" functions that indicate the probability they are satisfied, the two formulations correspond to two approaches of integrating reward (i.e. rationality in our model).
The stochastic transition formulation computes the expected rationality, and our formulation can be viewed as an approximation of maximum-likelihood estimated rationality -- we take $\max_\tau \lambda \log \Pr(\text{transitions in}~\tau~\text{are successful}) + Rationality(\bar{s},\bar{a},\tau)$. It would be an interesting extension to adopt the stochastic transition formulation (i.e., $\mathbb{E}_\tau \lambda \log \Pr(\text{transitions in}~\tau~\text{are successful}) + Rationality(\bar{s},\bar{a},\tau)$) and use a stochastic planner or MDP solver, although the planning time might be substantially increased.

Second, even if these two approaches behave differently in some cases, but it is unclear which one is better: this is a fundamental challenge in planning: how should the robot decide whether it has finished a task if there is no indication (such as rewards) from the environment?

\subsection{Discover and search with dependencies}
\label{app:ratskill-details:dependencies}

\paragraph{Evaluate goal classifiers when computing $\textit{first}(\bar{s}, o)$ for discovering dependencies.}
Recall that when discovering the dependencies, we need to compute $\textit{first}(\bar{s}, o)$, the smallest index $i$ such that $G_o(s_i)$ is true. We need to round $G_o(s_i)$ to Boolean value because we internally use the soft version of $G_o(\cdot)$.

For each subgoal $o$, we evaluate $G_o$ on all states in the training set to obtain the minimum and maximum output $\textit{MinG}_o$ and $\textit{MaxG}_o$, and we consider $G_o(s_i)$ to evaluate to true iff $G_o(s_i) \geq \sqrt{\textit{MinG}_o \cdot \textit{MaxG}_o}$.

\paragraph{Tuning hyperparameters.}
There are several hyper-parameters that need to be tuned: $\lambda, \alpha, \gamma, \beta$, and we discuss them separately below:
\begin{itemize}

\item $\lambda$ is determines the importance of satisfying the transition condition when transiting compared to regular action cost. In the environment, the cost of a single move $\mathcal{C}(s,a)$ is set to $0.1$, and $\lambda$ is set to $1$.

\item $\alpha$ is called the inverse rationality, the higher $\alpha$, the more rationality is assumed of the agent i.e. the agent is assume to take the optimal action with higher strategy. We set $\alpha=1$ in our experiments.

\item $\lambda, \alpha$  and the environmental action costs jointly determine the weight of FSM transitions, as well as the tolerance of suboptimality in the demonstrations (the higher $\alpha$ the lower tolerance). Our model is not sensitive to $\alpha$ (setting $0.1<\alpha<10$ have similar performance) because the trajectories in our dataset are close to optimal, and our model is a bit sensitive to the ratio of $\lambda$ to the action cost. We found that expected-number-of-steps-per-subgoal $\times$ action-cost is a good value for $\lambda$---it gives about the same weight to the transition appropriateness and the rationality along the path to achieve it.

\item $\gamma$ is the weight of the classification loss in the loss function $\mathcal{J}(\theta)$. We set $\gamma=0.1$ and we found that $0.01<\gamma<0.1$ all work well.

$\beta$ is to adjust the base for the softmax function for the classification, which is set to $1$. Note that $\beta$ and $\gamma$ jointly determine the weight of classification loss. Similarly, we have found that $0.1<\beta<1$ all work well.
\end{itemize}

\paragraph{Scalability and complexity of instruction generating.}

Meanwhile, the efficiency of our hierarchical search can be justified theoretically, even at the worst case when we are doing enumerating search approach without well-discovered dependencies. Say we have a subgoal set $\mathcal{O}$, a primitive action set $\mathcal A$, and each subgoal can be completed in $l$ actions, and the task can be achieved by sequencing $m$ subgoals.
Our two-level search generates up to $O(|\mathcal O|^m)$ candidate subgoal sequences and searching each sequence takes $O(m |\mathcal A|^l)$ time. Thus, the worst-case complexity is $O(m|\mathcal O|^m|\mathcal A|^l)$ which is still better than a pure primitive-level search, which is at worst $O(|\mathcal A|^{ml})$, because the number of subgoals $|\mathcal{O}|$ is usually much smaller than $|\mathcal A|^l$ (the number of all possible length $l$ sequences). %

\section{Dataset}
\label{app:dataset}
\subsection{\crafting}
Our \crafting environment is based on the Crafting environment introduced by \cite{Chen2020Ask}. The environment has a crafting agent that can move in a grid world, collect resources, and craft items.
In the original environment, every crafting rule is associated with a unique crafting station (\eg, paper must be crafted on the paper station). We modified the rules such that some crafting rules can share a common crafting station (\eg, both  arrows and swords can be crafted on a weapon station).
We add additional tiles: doors and rivers into the environment. Toggling a specific switch will open all doors. Otherwise, the agent can move across doors when they are holding a key. Meanwhile, the agent can move across rivers when they have a boat in their inventory.

We have used $47$ object types in \crafting including obstacles (\eg, river tiles, doors), items (\eg, axe), resources (\eg, trees), and crafting stations. We use $27$ rules for mining resources and crafting items. When the agent is at the same tile as another object, the {\it toggle} action will trigger the object-specific interaction. For item, the {\it toggle} action will pick up the object. For resource, the {\it toggle} action will mine the resource if the agent has space in their inventory and has the required tool for mining this type of resource (\eg, pickaxe is needed for mining iron ore).

\begin{table}[!t]
    \centering
    \small
    \begin{tabular}{p{0.2\columnwidth}p{0.2\columnwidth}p{0.2\columnwidth}p{0.2\columnwidth}}
    \toprule
      \multicolumn{4}{c}{Primitive} \\
      \midrule
      grab-pickaxe &  grab-axe & grab-key & toggle-switch \\
      craft-wood-plank & craft-stick & craft-shears & craft-bed \\
      craft-boat & craft-sword & craft-arrow & craft-cooked-potato \\
      craft-iron-ingot & craft-gold-ingot & craft-bowl & craft-beetroot-soup \\
      craft-paper & mine-gold-ore & mine-iron-ore & mine-sugar-cane \\
      mine-coal & mine-wood & mine-feather & mine-wool \\
      mine-potato & mine-beetroot & & \\

    \midrule
    \multicolumn{4}{c}{Compositional} \\
    \midrule

     \multicolumn{2}{l}{grab-pickaxe} & \multicolumn{2}{l}{grab-axe}  \\ \multicolumn{2}{l}{grab-key} & \multicolumn{2}{l}{toggle-switch} \\
     \multicolumn{2}{l}{mine-wood \dthen craft-wood-plank} & \multicolumn{2}{l}{craft-wood-plank \dthen craft-stick} \\
     \multicolumn{2}{l}{craft-iron-ingot \dor craft-gold-ingot \dthen craft-shears} &  \multicolumn{2}{l}{mine-wool \dand craft-wood-plank \dthen craft-bed} \\
     \multicolumn{2}{l}{craft-wood-plank \dthen craft-boat} & \multicolumn{2}{l}{craft-iron-ingot \dand craft-stick \dthen craft-sword} \\
     \multicolumn{2}{l}{mine-feather \dand craft-stick \dthen craft-arrow}
      & \multicolumn{2}{l}{mine-potato \dand mine-coal \dthen craft-cooked-potato} \\
      \multicolumn{2}{l}{mine-iron-ore \dand mine-coal \dthen craft-iron-ingot} & \multicolumn{2}{l}{mine-gold-ore \dand mine-coal \dthen craft-gold-ingot} \\
      \multicolumn{2}{l}{craft-wood-plank \dor craft-iron-ingot \dthen craft-bowl} & \multicolumn{2}{l}{craft-bowl \dand mine-beetroot \dthen craft-beetroot-soup} \\
      \multicolumn{2}{l}{mine-sugar-cane \dthen craft-paper} & \multicolumn{2}{l}{grab-pickaxe \dthen mine-gold-ore} \\
      \multicolumn{2}{l}{grab-pickaxe \dthen mine-iron-ore} & \multicolumn{2}{l}{grab-pickaxe \dor grab-axe \dthen mine-sugar-cane} \\
      \multicolumn{2}{l}{grab-pickaxe \dthen mine-coal} & \multicolumn{2}{l}{grab-axe \dthen mine-wood} \\
      \multicolumn{2}{l}{craft-sword \dthen mine-feather} & \multicolumn{2}{l}{craft-shears \dor craft-sword \dthen mine-wool} \\
      \multicolumn{2}{l}{grab-axe \dor mine-coal \dthen mine-potato} & \multicolumn{2}{l}{grab-axe \dor grab-pickaxe \dthen mine-beetroot} \\

    \midrule
    \multicolumn{4}{c}{Novel} \\
    \midrule
    \multicolumn{4}{l}{1. mine-sugar-cane \dthen craft-paper} \\
    \multicolumn{4}{l}{2. mine-potato \dand (gran pickaxe then mine-coal) \dand craft-cooked-potato} \\
    \multicolumn{4}{p{360pt}}{3. mine-beetroot \dand (grab-axe then mine-wood then craft-wood-plank then craft-bowl) \dthen craft-beetroot-soup} \\
    \multicolumn{4}{p{360pt}}{4. grab-axe \dthen mine-wood \dthen craft-wood-plank \dthen grab-pickaxe \dthen mine-iron-ore \dand mine-coal \dthen craft-iron-ingot \dthen craft-shears \dthen mine-wool \dthen craft-bed} \\
    \multicolumn{4}{p{360pt}}{5. grab-axe \dthen mine-wood \dthen craft-wood-plank \dthen craft-stick \dthen grab-pickaxe \dthen mine-iron-ore \dand mine-coal \dthen craft-iron-ingot \dthen craft-sword \dthen mine-feather \dthen mine-wood \dthen craft-wood-plank \dthen craft-stick \dthen craft-arrow} \\
    \multicolumn{4}{l}{6. grab-key \dthen grab-axe} \\
    \multicolumn{4}{l}{7. toggle-switch \dthen mine-beetroot} \\
    \multicolumn{4}{l}{8. grab-axe \dthen mine-wood \dthen craft-wood-plank \dthen craft-boat \dthen mine-sugar-cane} \\
    \multicolumn{4}{l}{9. grab-axe \dthen mine-wood \dthen craft-wood-plank \dthen craft-boat \dthen grab-pickaxe} \\
    \multicolumn{4}{p{360pt}}{10. grab-key \dthen grab-axe \dthen mine-wood \dthen craft-wood-plank \dthen craft-boat \dthen mine-potato} \\
    \multicolumn{4}{p{360pt}}{11. grab-key \dor (grab-axe \dthen mine-wood \dthen craft-wood-plank \dthen craft-boat) \dthen grab-pickaxe \dthen mine-gold-ore} \\
    \multicolumn{4}{p{360pt}}{12. grab-axe \dthen mine-wood \dthen craft-wood-plank \dthen craft-boat \dthen grab-key \dor toggle-switch \dthen grab-pickaxe \dthen mine-iron-ore \dand mine-coal \dthen craft-iron-ingot} \\
    \bottomrule
    \end{tabular}
    \vspace{5pt}
    \caption{Task descriptions in the {\it primitive}, {\it compositional} and {\it novel} sets for the \crafting.}
    \label{tab:supp:crafting-tasks}
\end{table}

\paragraph{State representation.} The state representation of \crafting consists of three parts.

\begin{enumerate}
    \item The {\it global feature} contains the size of grid world, the location of the agent, and the inventory size of the agent.
    \item The {\it inventory} feature contains an unordered list of objects in the agent's inventory. Each of them is represented as a one-hot vector indicating its object type.
    \item The {\it map} feature contains all objects on the map, including obstacles, items, resources, and crafting stations. Each of them is represented by a one-hot type encoding, the location (as integer values), and state (\eg, {\it open} or {\it closed} for doors).
\end{enumerate}

\paragraph{Action.} In \crafting, there are $5$ primitive level actions: {\it up}, {\it down}, {\it left}, {\it right}, and {\it toggle}. The first four actions will move the agent in the environment, while the {\it toggle} action will try to interact with the object in the same cell as the agent.

\paragraph{State feature extractor.}
Since our state representation contain a varying number of objects, we extract a vector representation of the environment with a relational neural network: Neural Logic Machines~\citep{dong2018neural}.

Concretely, we extract the inventory feature and the map feature separately. For each item in the inventory, we concatenate its input representation (\ie, the object type) with the global input feature. We process each item with the same fully-connected layer with ReLU activation. Following NLM~\cite{dong2018neural}, we use a max pooling operation to aggregate the feature for all inventory objects, resulting in a 128-dim vector. We use a similar architecture (but different neural network weights) to process all objects on the map. Finally, we concatenate the extracted inventory feature (128-dim), the map feature (128-dim), and the global feature (4-dim) as the holistic state representation. Thus, the output feature dimension for each state is 260.

\paragraph{Task definitions.}

\begin{table}[!t]
    \centering
    \begin{tabular}{lll}
    \toprule
      Final Goal & Steps & Example full instruction \\
      \midrule
      mine-wood & 2 & grab-axe \taskthen mine-wood \\
      craft-paper & 2 & mine-sugar-cane \taskthen craft-paper \\
      craft-beetroot-soup & 3 & (mine-beetroot \taskand craft-bowl) \taskthen craft-beetroot-soup \\
      craft-bed & 3 & (craft-wood-plank \taskand mine-wool) \taskthen craft-bed \\

      \multirow{2}{*}{craft-gold-ingot} & \multirow{2}{*}{4} & grab-pickaxe \taskthen (mine-gold-ore \taskand mine-coal)  \\
      & & \taskthen craft-gold-ingot \\

      \multirow{2}{*}{craft-boat} & \multirow{2}{*}{4} & grab-axe \taskthen mine-wood \taskthen craft-wood-plank  \\
      & & \taskthen craft-boat \\

      \multirow{2}{*}{craft-cooked-potato} & \multirow{2}{*}{4} & ((grab-pickaxe \taskthen mine-coal) \taskand mine-potato)  \\
       &   &  \taskthen craft-cooked-potato \\

      \multirow{2}{*}{craft-shears} & \multirow{2}{*}{5} & grab-pickaxe \taskthen (mine-coal \taskand mine-iron-ore)  \\
       &  &  \taskthen craft-iron-ingot \taskthen craft-shears \\
    \bottomrule
    \end{tabular}
    \vspace{5pt}
    \caption{Task descriptions used in search for a single final goal in \crafting.}
    \label{tab:supp:crafting-plan-search-tasks}
\end{table}

We list the task descriptions in the {\it primitive}, the {\it compositional}, and the {\it novel} splits \tbl{tab:supp:crafting-tasks}. \tbl{tab:supp:crafting-plan-search-tasks} lists 8 final goals and corresponding full instructions that are used in search for a single final goal.

\subsection{\playroom}

We build our \playroom environment following Konidaris~\etal~\cite{Konidaris2018Skills}. Specifically, we have added obstacles into the environment. The environment contains an agent, 6 effectors (a ball, a bell, a light switch, a button to turn on the music, a button to turn off the music and a monkey), and a fix number of obstacles. The agent and the effectors have fixed shapes. Thus, their geometry can be fully specified by their location and orientation. For simplicity, we have also fixed the shape and the location of the obstacles.

\paragraph{State representation.} We represent the pose of the agent by a 3D vector including the x, y coordinates (real-valued) and its rotation (real-valued, in $[-\pi, \pi)$. The state representation consist of the pose of the agent (as a 3-dimensional vector) and the locations of six effectors (as 6 2-dimensional vectors). Note that the state representation does not contain the shapes nor the locations of obstacles as they remain unchanged throughout the experiment. We concatenate these 7 vectors as the state representation.

\paragraph{Action.} The agent has a 3-dimensional action space: $[-1, 1]^3$. That is, for example, at each time step, the agent can at most move 1 meter along the x axis. We perform collision checking when the agent is trying to make a movement. If an action will result in a collision with objects or obstacles in the environment, the action will be treated as invalid and the state of the agent will not change.

\paragraph{Task definitions.}

We list the task descriptions in each of the {\it primitive}, {\it compositional} and {\it novel} set of the \playroom in Table \ref{tab:supp:playroom-tasks}

\begin{table}[ht]
    \centering
    \small
    \begin{tabular}{p{0.3\columnwidth}p{0.3\columnwidth}p{0.3\columnwidth}}
    \toprule
      \multicolumn{3}{c}{Primitive} \\
      \midrule
      play-with-ball & ring-bell & turn-on-light \\
      touch the mounkey & turn-off-music & turn-on-music \\

    \toprule
    \multicolumn{3}{c}{Compositional \textnormal{(designed meaningful tasks)}} \\
    \midrule
    \multicolumn{3}{l}{play-with-ball} \\ %
    \multicolumn{3}{l}{turn-on-light \taskthen ring-bell} \\ %
    \multicolumn{3}{l}{turn-on-music \taskand play-with-ball \taskthen touch the monkey} \\ %
    \multicolumn{3}{l}{play-with-ball \taskthen turn-on-light} \\ %
    \multicolumn{3}{l}{turn-on-music \taskand play-with-ball \taskthen turn-off-music} \\ %
    \multicolumn{3}{l}{turn-on-music \taskor play-with-ball} \\ %
    \multicolumn{3}{l}{turn-off-music \taskthen play-with-ball \taskthen turn-on-music} \\ %
    \multicolumn{3}{l}{turn-on-music \taskand play-with-ball \taskand turn-on-light \taskthen ring-bell} \\ %
    \toprule
    \multicolumn{3}{c}{Novel \textnormal{(randomly sampled)}} \\
    \midrule

    \multicolumn{3}{l}{play-with-ball \taskthen turn-on-light \taskor ring-bell} \\
    \multicolumn{3}{l}{turn-on-music \taskthen turn-on-light} \\
    \multicolumn{3}{l}{turn-on-music \taskthen turn-on-light} \\
    \multicolumn{3}{l}{play-with-ball \taskthen touch the monkey} \\
    \multicolumn{3}{l}{turn-on-music \taskthen turn-off-music} \\
    \multicolumn{3}{l}{turn-on-music \taskand ring-bell \taskthen touch the monkey} \\
    \multicolumn{3}{l}{ring-bell \taskthen touch the monkey \taskthen turn-on-light} \\
    \multicolumn{3}{l}{turn-on-light \taskand (ring-bell \taskor turn-on-music) \taskthen play-with-ball} \\
    \bottomrule
    \end{tabular}
    \vspace{5pt}
    \caption{Task descriptions in the {\it primitive}, {\it compositional} and {\it novel} sets for the \playroom.}
    \label{tab:supp:playroom-tasks}
\end{table}

\clearpage
\section{Baseline Implementation Details}
\label{app:implementations}

In this section, we present the implementation details of \model and other baselines. Without further notes, through out this section, we will be using the same LSTM encoder for task descriptions in $\tasklang$, and the same LSTM encoder for state sequences. The architecture of both encoders will be presented in \sectapp{app:implementation:lstm}.

\subsection{LSTM}
\label{app:implementation:lstm}
\paragraph{Task description encoder.} We use a bi-directional LSTM~\cite{hochreiter1997long} with a hidden dimension of 128 to encode the task description. The vocabulary contains all primitive subgoals, parentheses, and three connectives ({\it and}, {\it or}, and {\it then}). We perform an average pooling on the encoded feature for both directions, and concatenate them as the encoding for the task description. Thus, the output dimension is 256.

\paragraph{State sequence encoder.} For a given state sequence $\bar{s} = \{ s_i \}$, we first use a fully-connected layer to map each state $s_i$ into a 128-dimensional vector. Next, we feed the sequence into a bi-directional LSTM module. The hidden dimension of the LSTM is 128. We perform an average pooling on the encoded feature for both directions, and concatenate them as the encoding for the state sequence.

\paragraph{Training.} In our LSTM baseline for task recognition, we concatenate the state sequence feature and the task description feature, and use a 2-layer multi-layer perceptron (MLP) to compute the score of the input tuple: (trajectory, task description). The LSTM model is trained for 100 epochs on both environments. Each epoch contains 30 training batches that are randomly sampled from training data. The batch size is 32. We use the RMSProp optimizer with a learning rate decay from $10^{-3}$ to $10^{-5}$.

\subsection{Inverse Reinforcement Learning (IRL)}
The IRL baseline uses an LSTM model to encode task descriptions. We use different parameterizations for the reward function and the Q function in two datasets.

\paragraph{\crafting}
Since the task description may have complex temporal structures, the reward value does not only condition on the current state and but all historical states.
Therefore, instead of $Q(s,a | t)$ and $R(s,a,s' | t)$, we use $Q(\bar{s},a | t)$ and $R(\bar{s},a,s' | t)$ to parameterize the Q function and reward function, where $s$ is the current state, $a$ the action, $t$ the task description, $s'$ the next state, and $\bar{s}$ the historical state sequence from the initial state to the current state.

We use neural networks to approximate the Q function and reward function. For both of them, $\bar{s}$ is first encoded by an LSTM model into a fixed-length vector embedding. We simply concatenate the historical state encoding and the task description encoding, and then use a fully-connected layer to map the feature into a 5-dimensional vector. Each entry corresponds to the Q value or the reward value for a primitive action.

\paragraph{\playroom}

The Q function and reward function in \playroom also condition on all historical states. In \playroom, we parameterize the value of each state: $V(\bar{s})$, instead of $Q(\bar{s}, a)$. We parameterize $R(\bar{s},a,s')$ as $R(\bar{s}, s')$.

The input to our reward function network is composed of three parts: the vector encoding of the historical state sequence, the vector encoding for the next state $s'$, and the task description encoding. We concatenate all three vectors and run a fully-connected layer with a logSigmoid activation function.\footnote{We have experimented with no activation function, Sigmoid, and logSigmoid activations, and found that the logSigmoid activation works the best.}

In \playroom, since we do not directly parameterize the Q value for all actions in the continuous action space, in order to sample the best action at each state $s$ for plan execution, we first randomly sample 20 valid actions from the action space (\ie, actions that do not lead to collision), and choose the action that maximizes the Q function: $Q(\bar{s} \cup s', a)$, where $\bar{s}$ is the historical state seuqnce and $s'$ is the next state after taking $a$.

\paragraph{Value iteration.}
Both environments have a very large (\crafting) or even infinite (\playroom) state space. Thus it is impossible to run value iteration on the entire state space. Thus, at each iteration, for a given demonstration trajectory $(\bar{s}_e, \bar{a}_e)$, we construct a self exploration trajectory $(\bar{s}_p, \bar{a}_p)$ that share the same start state as $\bar{s}_e$~\footnote{Since running self-exploration in \playroom is too slow, in practice, we only generate self-exploration trajectories for 4 trajectories in the input batch.}. We run value iteration on $\{\bar{s}_e\} \cup \{\bar{s}_p\}$. For states not in this set, we use the Q function network to approximate their values.

\paragraph{Training.}
For both \crafting and \playroom, we train the IRL model for 60 epochs. We set the batch size to be 32 and each epoch has 30 training batches. We use a replay buffer that can store 100,000 trajectories. For both environments, we use the Adam optimizer with a learning rate decay from $10^{-3}$ to $10^{-5}$. We have found the IRL method unstable to train in the \playroom environment. Thus, in \playroom, we use a warm-up training procedure. In the first 18(30\%) epochs, we set $\gamma=0$ for a ``warm start'', and for rest of the epochs we use $\gamma=0.5$, where $\gamma$ is the discount factor in the Q function.

\subsection{Behavior Cloning (BC)}
BC learns a policy $\pi(\bar{s}, a | t)$ from data, where $t$ is the task description, $a$ a primitive action, and $\bar{s}$ the historical state sequence. The state sequence $\bar{s}$ is first encoded by an LSTM model into a fixed-length vector embedding.

In \crafting, we use a fully-connected layer with softmax activation to parameterize $\pi(a | \bar{s}, t)$. Specifically, the input to the fully-connected layer is the concatenation of the vector encoding of $\bar{s}$ and the vector encoding of the task description $t$.

In \playroom, we use two fully-connected (FC) layers to parameterize $\pi(a | \bar{s}, t)$. Specifically, we parameterize $\pi(a | \bar{s}, t)$ as a Gaussian distribution. The first FC layer has a Tanh activation and parameterizes the mean $\mu$ of the Gaussian. The second FC layer has a Sigmoid activation and paramerizes the standard variance $\sigma^2$ of the Gaussian.

To make this model more consistent with our \bcfsm model, in both environments, we also train a module to compute the termination condition of the trajectory. That is, a neural network that maps $\bar{s}$ to a real value in $[0, 1]$, indicating the probability of terminating the execution. Denote the output of this network as $\textit{stop}(\bar{s})$. At each time step, the agent will terminate its policy with probability $\textit{stop}(\bar{s})$. We modulate the probability for other actions $a$ as $\pi(a | \bar{s}, t) \cdot (1 - \textit{stop}(\bar{s}))$

For planning in \crafting, at each step, we choose the action with the maximum probability (including the option to ``terminate'' the execution). In \playroom, we always take the ``mean'' action parameterized by $\pi(a | \bar{s}, t)$ until we reach the maximum allowed steps.

We then define the score of a task given a trajectory, $\textit{score}(\bar{s}, \bar{a}, t)$, as the sum of log-probabilities of the actions taken at each step. We train this model with the same loss and training procedure as \model. We train the model for 100 epochs using the Adam optimizer with a learning rate decay from $10^{-3}$ to $10^{-5}$.

\subsection{Behavior Cloning with \fsm (\bcfsm)}

\bcfsm represents task description as an \fsm, in the same way as our model \model. It represents each subgoal $o$ as a tuple: $\langle \pi_o(s a), \textit{stop}_o(s) \rangle$, corresponding to the subgoal-conditioned policy and the termination condition.

\paragraph{Task recognition.}
The task recognition procedure for \bcfsm jointly segments the trajectory and computes the consistency score between the task description and the input trajectory. In particular, our algorithm will assign an \fsm state $v_i$ to each state $s_i$, and insert several action. We use a dynamic programming algorithm (similar to the one used by our algorithm for \model) to find the assignment that maximize the overall score:
\[
\textit{score}(\bar{s}, \bar{v}, \bar{a}) \coloneqq \prod_{i} p(a_i | s_i, v_i, t)
\]
\[
p(a_i | s_i, v_i, t) = \left\{\begin{array}{ll}
   \pi(a_i|s_i,v_i)\cdot (1-\textit{stop}(s_i,v_i))  & \text{if $a \in \mathcal {A}$ is a primitive action} \\
   \textit{stop}(s_i,v_i)  & \text{if $a \in E_t$ is an \fsm transition}
\end{array}\right.
\]

\paragraph{Planning.}
We use the same strategy as the basic Behavior Cloning model to choose actions at each step, conditioned on the current \fsm state. \bcfsm handles branches in the \fsm in the same way as our algorithm for \model.

\subsection{Computational Source Used for Training and Testing}

\begin{table}[!t]
    \centering
    \begin{tabular}{lccccc}
    \toprule
    \multirow{2}{*}{Model}  & \multirow{2}{*}{\#Epochs} & \multicolumn{2}{c}{Crafting World} & \multicolumn{2}{c}{Playroom} \\
      \cmidrule{3-6}
       &  & \mycellc{Training Time\\(minute/epoch)} & \mycellc{Planning Time\\(second/sample)} & \mycellc{Training Time\\(minute/epoch)} & \mycellc{Planning Time\\(second/sample)} \\
      \midrule
    RSG & 60 & 17.0&  4.6962 & 32.4 &  0.4313  \\
    LSTM & 200 & 0.4 &  N/A &  0.2 &  N/A \\
    IRL & 60 & 34.8&  0.7269  & 7.1 & 23.0187  \\
    BC & 150 & 0.3 &  0.1615  & 0.4 &  0.1688 \\
    BC(FSM) & 150 & 0.5 &  0.5423 & 0.5 &  0.1875 \\
    \bottomrule
    \end{tabular}
    \vspace{5pt}
    \caption{Training time per epoch and planning time per sample for all models.}
    \label{tab:comp-time}
\end{table}

In Table \ref{tab:comp-time}, we list the training time per epoch for our model and all baselines, and averaging time cost for each planning task when testing. All models are trained until convergence, and we list the number of epochs after which the loss stops going down.

\subsection{Generative Adversarial Imitation Learning}

We have also tested the Generative Adversarial Imitation Learning (GAIL)\citep{ho2016generative} as a baseline on the planning task on seen instructions. Since GAIL does not natually generalize to structured subgoals (FSMs), we used an LSTM to encode the task description as in the BC and IRL baselines. On CraftingWorld, GAIL achieves 19.2\% planning success rate on the Comp. split, and 1.0\% success rate on the Novel split. GAIL has a better performance compared to BC and BC-FSM, which we think is because it explores the environment during training. There is still a large performance gap between GAIL and RSG, mainly because GAIL does not have goal representations that helps compositional generalization. %

\end{document}